\documentclass[10pt,twocolumn,letterpaper]{article}

\usepackage{cvpr}
\usepackage{times}
\usepackage{epsfig}
\usepackage{graphicx}
\usepackage{amsmath}
\usepackage{amssymb}
\usepackage{breqn}
\usepackage{bbm}
\usepackage{caption}
\usepackage{subcaption}
\usepackage{color}
\usepackage{booktabs}
\usepackage{multirow}
\usepackage[lined,boxed, commentsnumbered]{algorithm2e}
\usepackage{nicefrac}

\usepackage[pagebackref=true,breaklinks=true,letterpaper=true,colorlinks,bookmarks=false]{hyperref}

\newcommand{\tj}{\tilde{J}}

\cvprfinalcopy 

\def\cvprPaperID{1739} 

\ifcvprfinal\fi
\begin{document}

\title{Deep Metric Learning via Lifted Structured Feature Embedding}

\author{Hyun Oh Song\\
Stanford University\\
\hspace{-1.5em} {\tt\small hsong@cs.stanford.edu} \hspace{-1.7em}
\and
Yu Xiang\\
Stanford University\\
{\tt\small yuxiang@cs.stanford.edu} \hspace{-1.5em}
\and
Stefanie Jegelka\\
MIT\\
{\tt\small stefje@csail.mit.edu} \hspace{-1.5em}
\and
Silvio Savarese\\
Stanford University\\
{\tt\small ssilvio@stanford.edu}
}

\maketitle

\begin{abstract}
Learning the distance metric between pairs of examples is of great importance for learning and visual recognition. With the remarkable success from the state of the art convolutional neural networks, recent works \cite{seanbell, facenet} have shown promising results on discriminatively training the networks to learn semantic feature embeddings where similar examples are mapped close to each other and dissimilar examples are mapped farther apart. In this paper, we describe an algorithm for taking full advantage of the training batches in the neural network training by lifting the vector of pairwise distances within the batch to the matrix of pairwise distances. This step enables the algorithm to learn the state of the art feature embedding by optimizing a novel structured prediction objective on the lifted problem. Additionally, we collected Online Products dataset: 120k images of 23k classes of online products for metric learning. Our experiments on the CUB-200-2011 \cite{cub}, CARS196 \cite{cars}, and Online Products datasets demonstrate significant improvement over existing deep feature embedding methods on all experimented embedding sizes with the GoogLeNet \cite{googlenet} network.
\end{abstract}

\section{Introduction}
\begin{figure}[thbp]
\centering
\includegraphics[width=0.48\textwidth]{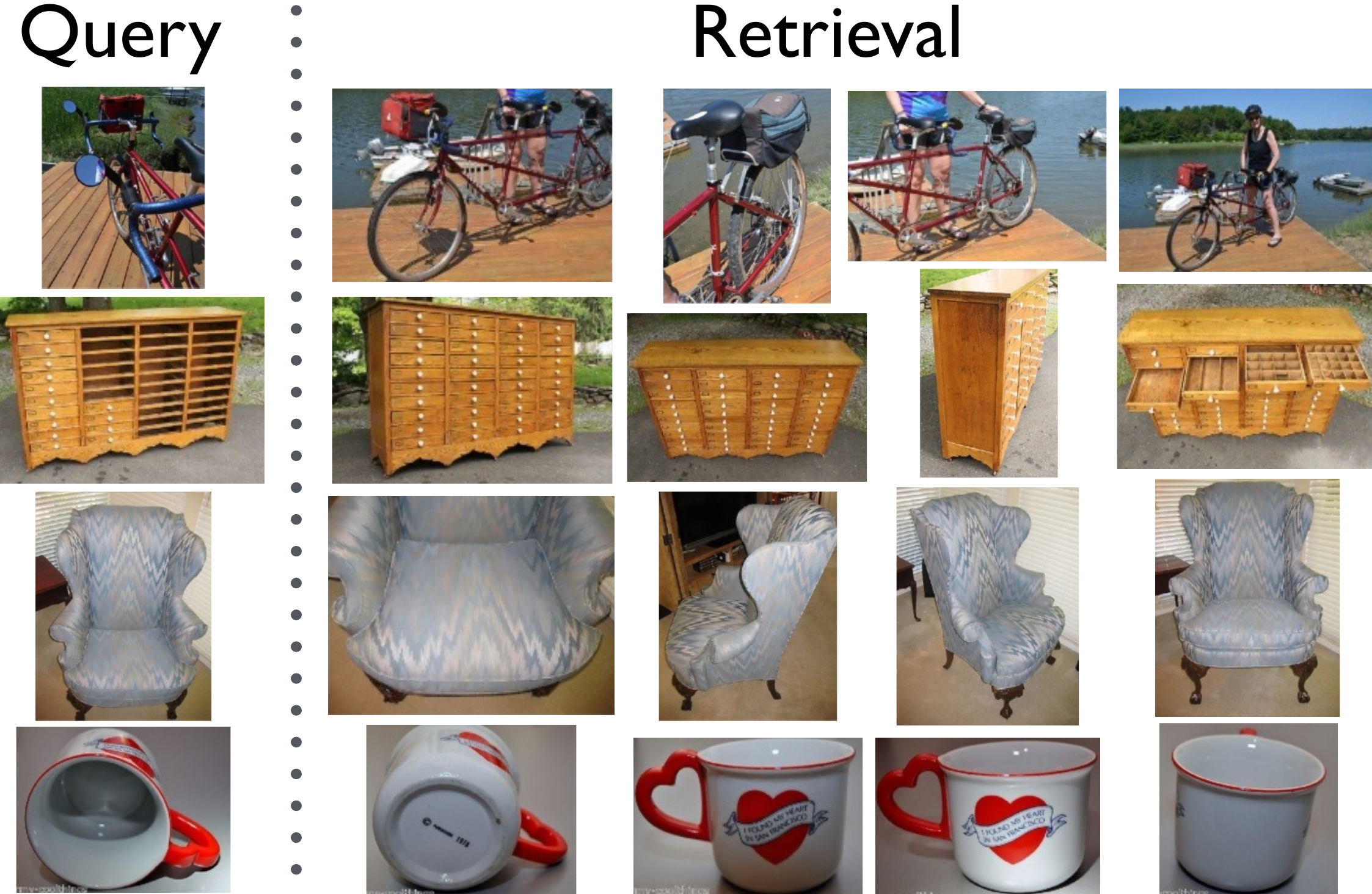}\\
\caption{Example retrieval results on our \emph{Online Products} dataset using the proposed embedding. The images in the first column are the query images.}
\label{fig:figure1}
\end{figure}

Comparing and measuring similarities between pairs of examples is a core requirement for learning and visual competence. Being able to first measure how similar a given pair of examples are makes the following learning problems a lot simpler. Given such a similarity function, classification tasks could be simply reduced to the nearest neighbor problem with the given similarity measure, and clustering tasks would be made easier given the similarity matrix. In this regard, metric learning \cite{nca, triplet, taylor} and dimensionality reduction \cite{pca, mds, lle, outofsample} techniques aim at learning semantic distance measures and embeddings such that similar input objects are mapped to nearby points on a manifold and dissimilar objects are mapped apart from each other. 

Furthermore, the problem of \emph{extreme classification} \cite{extreme_langford, extreme_varma} with enormous number of categories has recently attracted a lot of attention in the learning community. In this setting, two major problems arise which renders conventional classification approaches practically obsolete. First, algorithms with the learning and inference complexity linear in the number of classes become impractical. Second, the availability of training data per class becomes very scarce. In contrast to conventional classification approaches, metric learning becomes a very appealing technique in this regime because of its ability to learn the general concept of distance metrics (as opposed to category specific concepts) and its compatibility with efficient nearest neighbor inference on the learned metric space.

With the remarkable success from the state of the art convolutional neural networks \cite{alexnet, googlenet}, recent works \cite{seanbell, facenet} discriminatively train neural network to directly learn the the non-linear mapping function from the input image to a lower dimensional embedding given the input label annotations. In high level, these embeddings are optimized to pull examples with different class labels apart from each other and push examples from the same classes closer to each other. One of the main advantages of these discriminatively trained network models is that the network jointly learns the feature representation and semantically meaningful embeddings which are robust against intra-class variations.

However, the existing approaches \cite{seanbell, facenet} cannot take full advantage of the training batches used during the mini batch stochastic gradient descent training of the networks \cite{alexnet, googlenet}. The existing approaches first take randomly sampled pairs or triplets to construct the training batches and compute the loss on the individual pairs or triplets within the batch. Our proposed method \emph{lifts} the  \emph{vector} of pairwise distances ($O(m)$) within the batch to the \emph{matrix} of pairwise distances ($O(m^2)$).  Then we design a novel structured loss objective on the lifted problem. Our experiments show that the proposed method of learning the embedding with the structured loss objective on the lifted problem significantly outperforms existing methods on all the experimented embedding dimensions with the GoogLeNet \cite{googlenet} network.

We evaluate our methods on the CUB200-2011 \cite{cub}, CARS196 \cite{cars}, and \emph{Online Products} dataset we collected. The \emph{Online Products} has approximately $120k$ images and $23k$ classes of product photos from online e-commerce websites. To the best of our knowledge, the dataset is one of the largest publicly available dataset in terms of the number and the variety of classes. We plan to maintain and grow the dataset for the research community.

In similar spirit of general metric learning where the task is to learn a generic concept of similarity/distance, we construct our train and test split such that there is no intersection between the set of classes used for training versus testing. We show that the clustering quality (in terms of standard $\text{F}_1$ and NMI metrics \cite{manningbook}) and retrieval quality (in terms of standard Recall@K score) on images from previously unseen classes are significantly better when using the proposed embedding. Figure \ref{fig:figure1} shows some example retrieval results with the \emph{Online Products} dataset using the proposed embedding. Although we experiment on clustering and retrieval tasks, the conceptual contribution of this paper - lifting a batch of examples into a dense pairwise matrix and defining a structured learning problem - is generally applicable to a variety of learning and recognition tasks where feature embedding is employed. 
\section{Related works}

Our work is related to three lines of active research: (1) Deep metric learning for recognition, (2) Deep feature embedding with convolutional neural networks, and (3) Zero shot learning and ranking.\vspace{0.3em}

\noindent\textbf{Deep metric learning:}~  Bromley \emph{et al}. \cite{signatureVerification} paved the way on deep metric learning and trained Siamese networks for signature verification. Chopra \emph{et al}. \cite{faceVerification} trained the network discriminatively for face verification. Chechik \emph{et al}. \cite{triplet_ranking} learn ranking function using triplet \cite{triplet} loss. Qian \emph{et al}. \cite{nec} uses precomputed \cite{alexnet} activation features and learns a feature embedding via distance metric for classification. \vspace{0.5em}

\noindent\textbf{Deep feature embedding with state of the art convolutional neural networks:}~  Bell \emph{et al}. \cite{seanbell} learn embedding for visual search in interior design using contrastive \cite{contrastive} embedding, FaceNet \cite{facenet} uses triplet \cite{triplet} embedding to learn embedding on faces for face verification and recognition. Li \emph{et al}. \cite{chairs} learn a joint embedding shared by both 3D shapes and 2D images of objects. In contrast to the existing approaches above, our method computes a novel structured loss and the gradient on the lifted dense pairwise distance matrix to take full advantage of batches in SGD.\vspace{0.2em}

\noindent\textbf{Zero shot learning and ranking:}~  Frome \emph{et al}., Socher \emph{et al}., and Weston \emph{et al}. \cite{devise, socher, wasabi} leverage text data to train visual ranking models and to constrain the visual predictions for zero shot learning. Wang \emph{et al}. \cite{triplet_cvpr14} learns to rank input triplet of data given human rater's rank ratings on each triplets and also released a triplet ranking dataset with 5,033 triplet examples \cite{triplet_data}. However, the approach is not scalable with the size of the training data because it's very costly to obtain ranking annotations in contrast to multiclass labels (i.e., product name) and because the approach is limited to ranking the data in triplet form. Lampert \emph{et al}. \cite{lampert} does zero shot learning but with attributes (such as objects's color or shape) provided for both the train and the test data. On a related note, \cite{mensink, palatucci, rohrbach} do zero-shot learning for visual recognition but rely on the WordNet hierarchy for semantic information of the labels. \vspace{0.2em}

\noindent The paper is organized as follows. In section \ref{sec:review}, we start with a brief review of recent state of the art deep learning based embedding methods \cite{contrastive, facenet}. In section \ref{sec:method}, we describe how we lift the problem and define a novel structured loss. In section \ref{sec:implementation} and \ref{sec:evaluation}, we describe the implementation details and the evaluation metrics. We present the experimental results and visualizations in section \ref{sec:experiments}.
\section{Review}
\label{sec:review}
In this section, we briefly review recent works on discriminatively training neural networks to learn semantic embeddings. \vspace{0.5em}

\noindent \textbf{Contrastive embedding} \cite{contrastive} is trained on the paired data $\left\{\left(\mathbf{x}_i, \mathbf{x}_j, y_{ij}\right)\right\}$. Intuitively, the contrastive training minimizes the distance between a pair of examples with the same class label and penalizes the negative pair distances for being smaller than the margin parameter $\alpha$. Concretely, the cost function is defined as, 

\begin{align}
J =  \frac{1}{m}\sum_{(i,j)}^{m/2} y_{i,j} D_{i,j}^2 + (1-y_{i,j}) \left[\alpha - D_{i,j}\right]_{+}^2,
\end{align}

\noindent where $m$ stands for the number of images in the batch, $f(\cdot)$ is the feature embedding output from the network, $D_{i,j} = ||f(\mathbf{x}_i) - f(\mathbf{x}_j)||_2$, and  the label $y_{i,j} \in \{0,1\}$ indicates whether a pair $\left(\mathbf{x}_i, \mathbf{x}_j\right)$ is from the same class or not. The $[\cdot]_+$ operation indicates the hinge function $\max(0, \cdot)$. We direct the interested readers to refer \cite{contrastive, seanbell} for the details. \vspace{0.7em}

\noindent \textbf{Triplet embedding} \cite{triplet, facenet} is trained on the triplet data $\left\{\left( \mathbf{x}_a^{(i)}, \mathbf{x}_p^{(i)}, \mathbf{x}_n^{(i)} \right)\right\}$ where $\left(\mathbf{x}_a^{(i)}, \mathbf{x}_p^{(i)}\right)$ have the same class labels and $\left(\mathbf{x}_a^{(i)}, \mathbf{x}_n^{(i)}\right)$ have different class labels. The $\mathbf{x}_a^{(i)}$ term is referred to as an \emph{anchor} of a triplet. Intuitively, the training process  encourages the network to find an embedding where the distance between $\mathbf{x}_a^{(i)}$ and $\mathbf{x}_n^{(i)}$ is larger than the distance between $\mathbf{x}_a^{(i)}$ and $\mathbf{x}_p^{(i)}$ plus the margin parameter $\alpha$. The cost function is defined as,

\begin{align}
\begin{centering}
J = \frac{3}{2m}\sum_i^{m/3} \left[D_{ia,ip}^2 - D_{ia,in}^2 + \alpha\right]_{+}, 
\end{centering}
\label{eqn:triplet}
\end{align}

\noindent where $D_{ia,ip} = ||f(\mathbf{x}_i^a) - f(\mathbf{x}_i^p)||$ and $D_{ia,in} = ||f(\mathbf{x}_i^a) - f(\mathbf{x}_i^n)||$. Please refer to \cite{facenet, triplet} for the complete details.



\begin{figure}[thbp]
\centering
\begin{subfigure}[b]{0.4\textwidth}
\includegraphics[width=\textwidth]{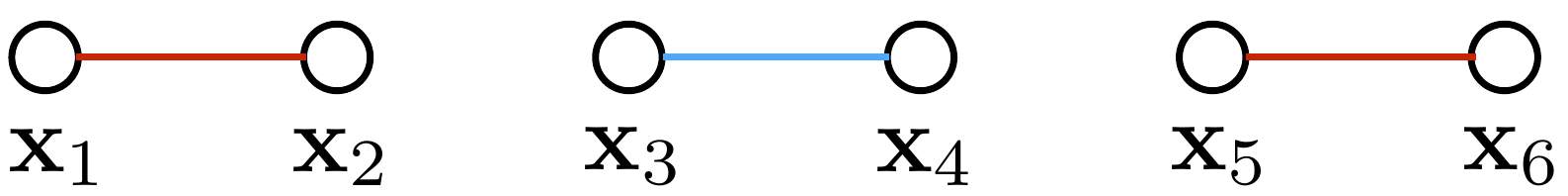}
\caption{Contrastive embedding}
\label{fig:contrastive}
\end{subfigure}\\ \vspace{0.3em}
\begin{subfigure}[b]{0.4\textwidth}
\includegraphics[width=\textwidth]{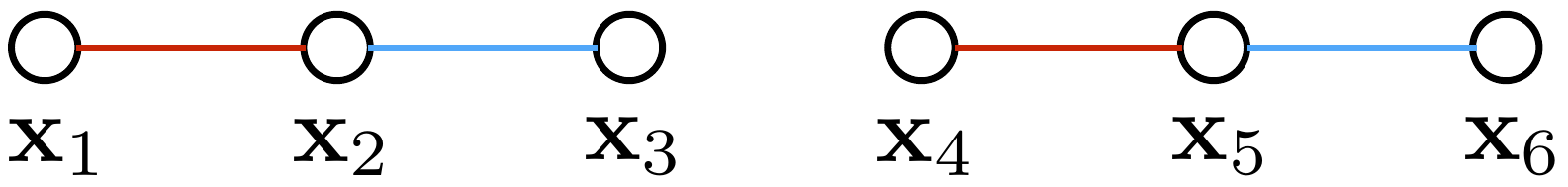}
\caption{Triplet embedding}
\label{fig:triplet}
\end{subfigure}\\ \vspace{0.3em}
\begin{subfigure}[b]{0.4\textwidth}
\includegraphics[width=\textwidth]{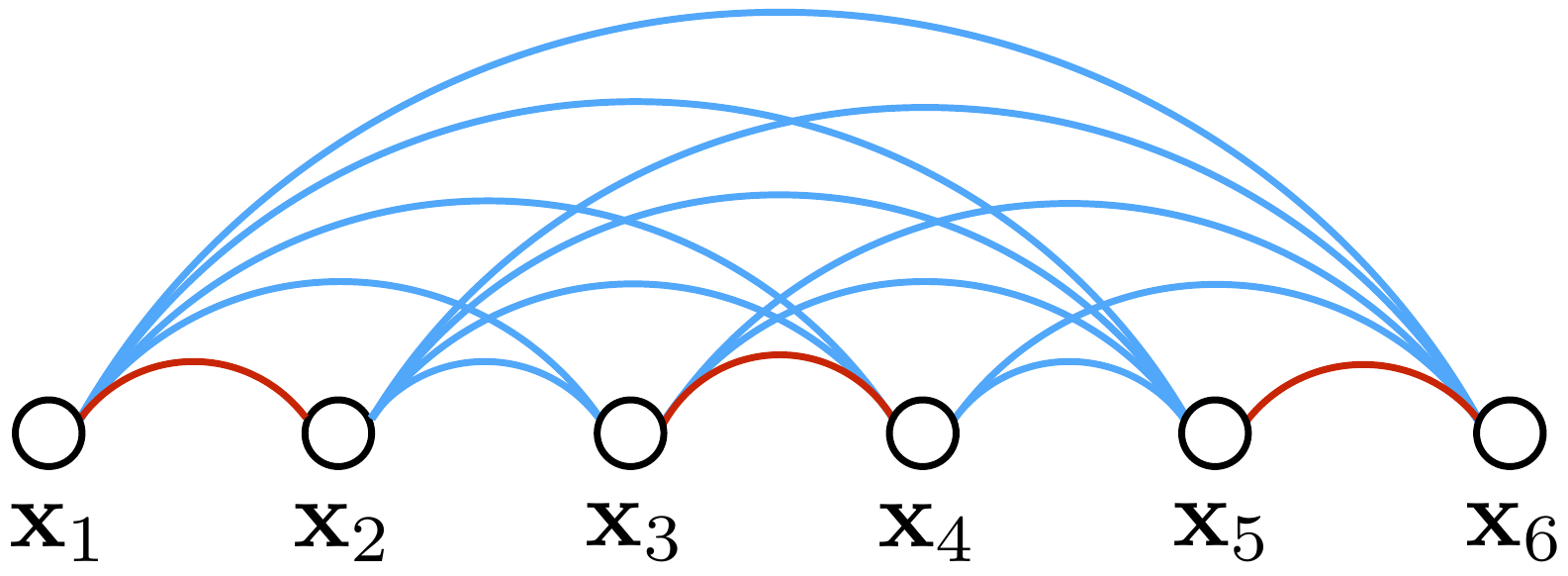}
\caption{Lifted structured embedding}
\label{fig:structsim}
\end{subfigure}
\caption{Illustration for a training batch with six examples. Red edges and blue edges represent similar and dissimilar examples respectively. In contrast, our method explicitly takes into account all pair wise edges within the batch.}
\label{fig:losses}
\end{figure} 

\section{Deep metric learning via lifted structured feature embedding}
\label{sec:method}

\begin{figure}[thbp]
\centering
\begin{subfigure}[b]{0.4\textwidth}
\includegraphics[width=\textwidth]{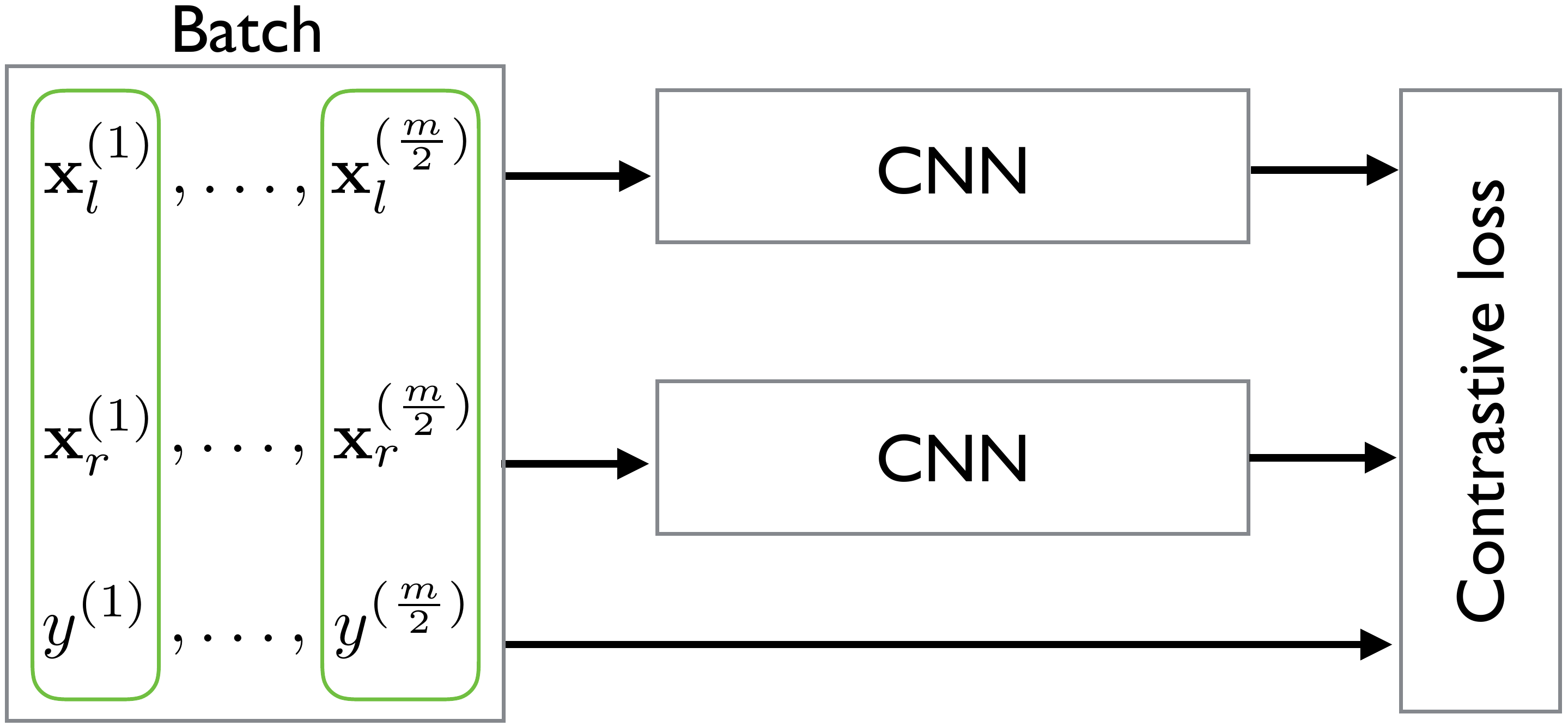}
\caption{Training network with contrastive embedding \cite{contrastive}}
\label{fig:contrastive_net}
\end{subfigure}\\ \vspace{0.2em}
\begin{subfigure}[b]{0.4\textwidth}
\includegraphics[width=\textwidth]{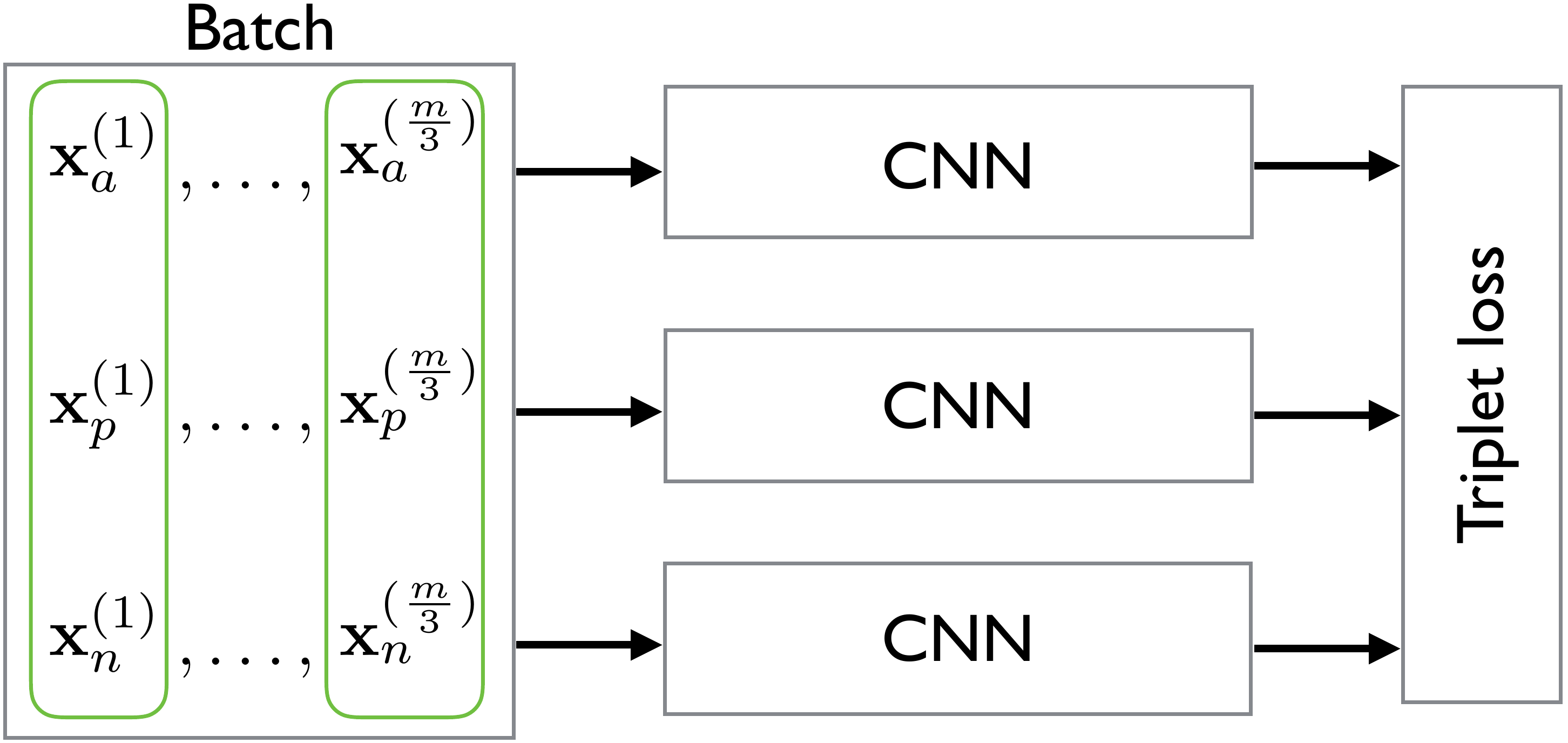}
\caption{Training network with triplet embedding \cite{triplet, facenet}}
\label{fig:triplet_net}
\end{subfigure}\\ \vspace{0.2em}
\begin{subfigure}[b]{0.38\textwidth}
\hspace{0.43em}\includegraphics[width=\textwidth]{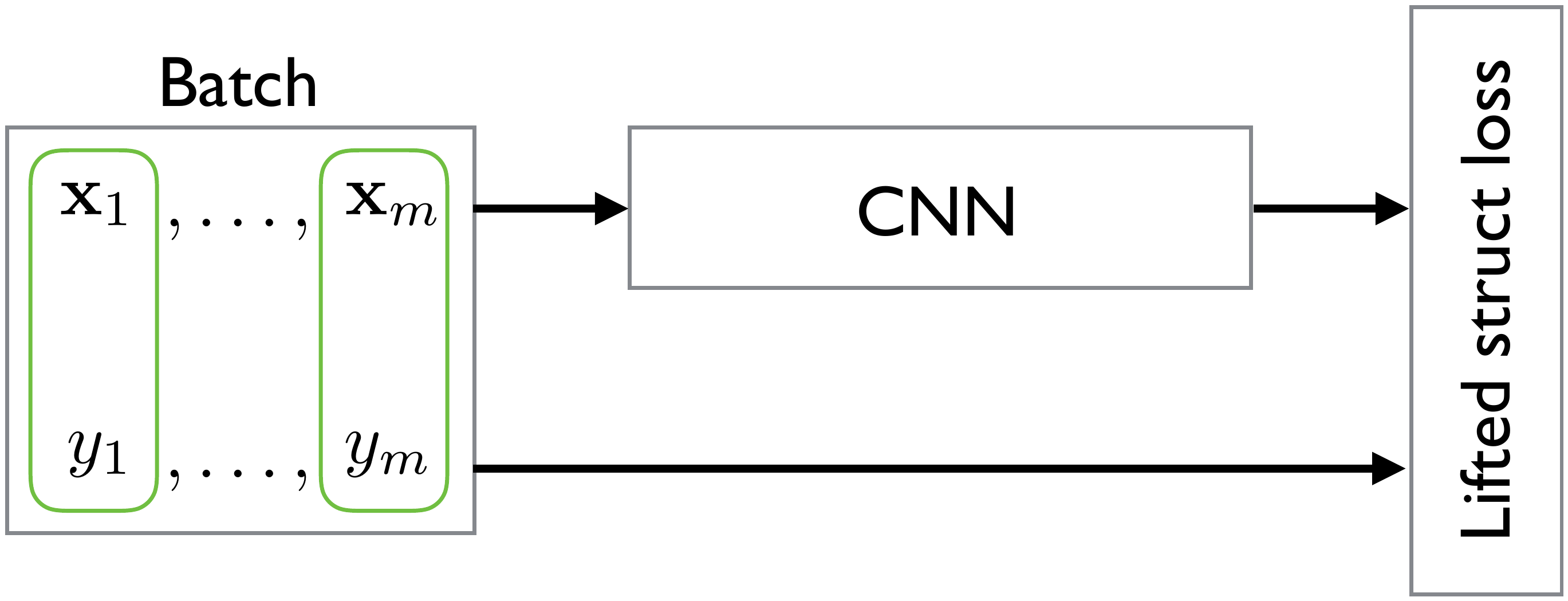}
\caption{Training network with lifted structure embedding}
\label{fig:structsim_net}
\end{subfigure}
\caption{Illustration for training networks with different embeddings. $m$ denotes the number of images in the batch. The green round box indicates one example within the batch. The network (a) takes as input binary labels, network (b) does not take any label input because the ordering of anchor, positive, negative encodes the label. The proposed network (c) takes as input the multiclass labels.}
\label{fig:networks}
\end{figure}

We define a structured loss function based on all positive and negative pairs of samples in the training set: 
{\small
\begin{align}
J = &\frac{1}{2 |\widehat{\mathcal{P}}|} \sum_{\left(i,j\right) \in \widehat{\mathcal{P}}} \max\left(0, ~J_{i,j}\right)^2,\\
J_{i,j} = &\max\left(\max_{\left(i,k\right) \in \widehat{\mathcal{N}}} \alpha - D_{i,k}, \max_{\left(j,l\right) \in \widehat{\mathcal{N}}} \alpha - D_{j,l}\right) + D_{i,j}\notag
\label{eqn:structsim}
\end{align}}
\noindent where $\widehat{\mathcal{P}}$ is the set of positive pairs and $\widehat{\mathcal{N}}$ is the set of negative pairs in the training set. This function poses two computational challenges: (1) it is non-smooth, and (2) both evaluating it and computing the subgradient requires mining all pairs of examples several times.

We address these challenges in two ways: First, we optimize a smooth upper bound on the function instead. Second, as is common for large data sets, we use a stochastic approach. However, while previous work implements a stochastic gradient descent by drawing pairs or triplets of points uniformly at random \cite{contrastive, seanbell, chairs}, our approach deviates from those methods in two ways: (1) it biases the sample towards including ``difficult'' pairs, just like a subgradient of $J_{i,j}$ would use the close negative pairs \footnote{Strictly speaking, this would be a subgradient replacing the nested max by a plus.}; (2) it makes use of the full information of the mini-batch that is sampled at a time, and not only the individual pairs.\\

Figures \ref{fig:contrastive} and \ref{fig:triplet} illustrate a sample batch of size $m=6$ for the contrastive and triplet embedding. Red edges in the illustration represent positive pairs (same class) and the blue edges represent negative pairs (different class) in the batch. In this illustration, it is important to note that adding extra vertices to the graph is a lot more costly than adding extra edges because adding vertices to the graph incurs extra I/O time and/or storage overhead.\\

To make full use of the batch, one key idea is to enhance the mini-batch optimization to use all $O(m^2)$ pairs in the batch, instead of $O(m)$ separate pairs. Figure \ref{fig:structsim} illustrates the concept of of transforming a training batch of examples to a fully connected dense matrix of pairwise distances. Given a batch of $c$-dimensional embedded features $X \in \mathbb{R}^{m \times c}$ and the column vector of squared norm of individual batch elements $\tilde{\mathbf{x}} = \left[ ||f(\mathbf{x}_1)||_2^2, \ldots, ||f(\mathbf{x}_m)||_2^2\right]^\intercal$, the dense pairwise squared distance matrix can be efficiently constructed by computing, $D^2 = \tilde{\mathbf{x}} \mathbf{1}^\intercal + \mathbf{1} \tilde{\mathbf{x}}^\intercal - 2 XX^\intercal$, where $D_{ij}^2 = || f(\mathbf{x}_i) - f(\mathbf{x}_j)||_2^2$. However, it is important to note that the negative edges induced between randomly sampled pairs carry limited information. Most likely, they are different from the much sharper, close (``difficult'') neighbors that a full subgradient method would focus on.\\

\begin{figure}[thbp]
\centering
\includegraphics[width=0.4\textwidth]{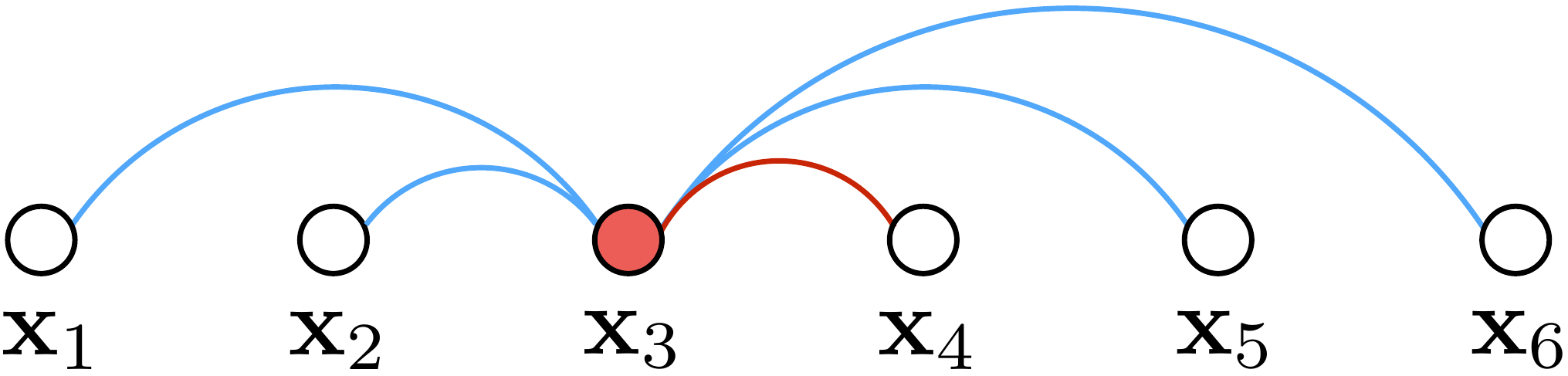}\\
\includegraphics[width=0.4\textwidth]{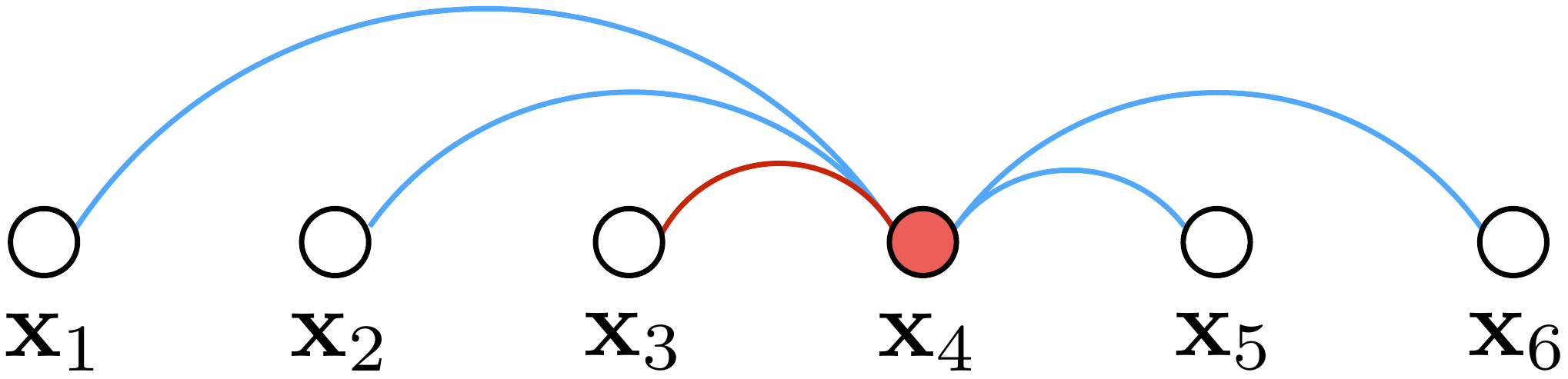}
\caption{Hard negative edge is mined with respect to each left and right example per each positive pairs. In this illustration with 6 examples in the batch, both $\mathbf{x}_3$ and $\mathbf{x}_4$ independently compares against all other negative edges and mines the hardest negative edge.}
\label{fig:mining}
\end{figure}

Hence, we change our batch to be not completely random, but integrate elements of importance sampling. We sample a few positive pairs at random, and then actively add their difficult neighbors to the training mini-batch. This augmentation adds relevant information that a subgradient would use. Figure \ref{fig:mining} illustrates the mining process for one positive pair in the batch, where for each image in a positive pair we find its close (hard) negative images. Note that our method allows mining the hard negatives from both the left and right image of a pair in contrast to the rigid triplet structure \cite{facenet} where the negative is defined only with respect to the predefined anchor point. Indeed, the procedure of mining hard negative edges is equivalent to computing the loss augmented inference in structured prediction setting \cite{Tsochantaridis/etal/04, Joachims/etal/09a}. Our loss augmented inference can be efficiently processed by first precomputing the pairwise batch squared distance matrix $D^2$. Figure \ref{fig:networks} presents the comprehensive visual comparison of different training structures (i.e. batch, label, network layout) with different loss functions. In contrast to other approaches (Fig. \ref{fig:contrastive_net} and \ref{fig:triplet_net}), our method greatly simplifies the network structure (Fig. \ref{fig:structsim_net}) and requires only one branch of the CNN.

Furthermore, mining the single hardest negative with nested $\max$ functions (eqn. \ref{eqn:structsim}) in practice causes the network to converge to a bad local optimum. Hence we optimize the following smooth upper bound $\tj(D(f(\mathbf{x})))$. Concretely, our loss function per each batch is defined as, 

{\small
\begin{align}
\hspace{-0.5em}\tj_{i,j} = &\log \left(\sum_{\left(i,k\right) \in \mathcal{N}} \exp\{\alpha - D_{i,k}\} + \sum_{\left(j,l\right) \in \mathcal{N}} \exp\{\alpha - D_{j,l}\} \right) + D_{i,j}\notag\\
\tilde{J} = &\frac{1}{2 |\mathcal{P}|} \sum_{\left(i,j\right) \in \mathcal{P}} \max\left(0, ~\tj_{i,j}\right)^2,
\end{align}}


\begin{figure*}[thbp]
\centering
\begin{subfigure}[b]{0.3\textwidth}
\includegraphics[width=\textwidth]{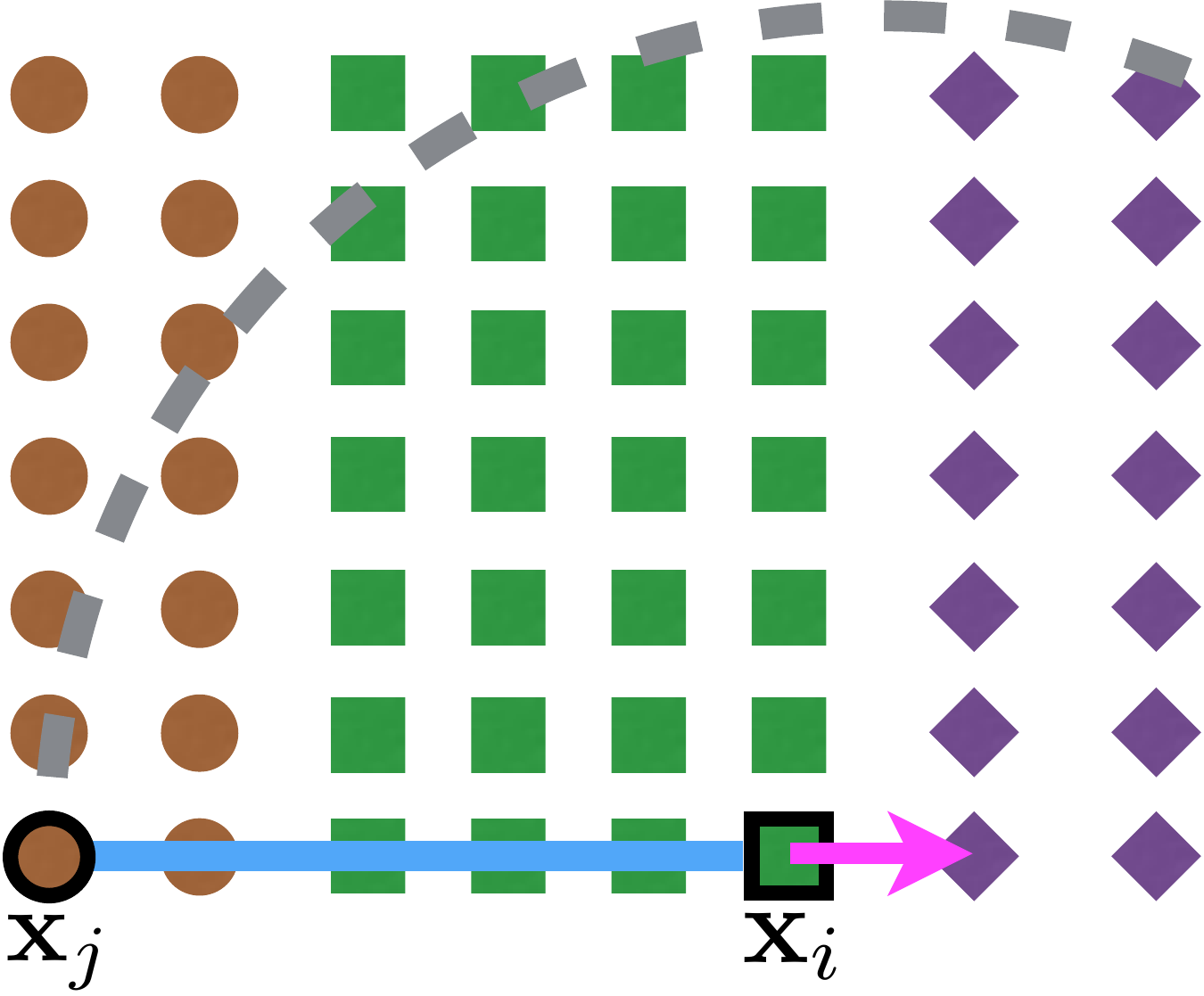} 
\caption{Contrastive embedding}
\label{fig:losses_contrastive}
\end{subfigure}
\hspace{0.02\textwidth}
\begin{subfigure}[b]{0.31\textwidth}
\includegraphics[width=\textwidth]{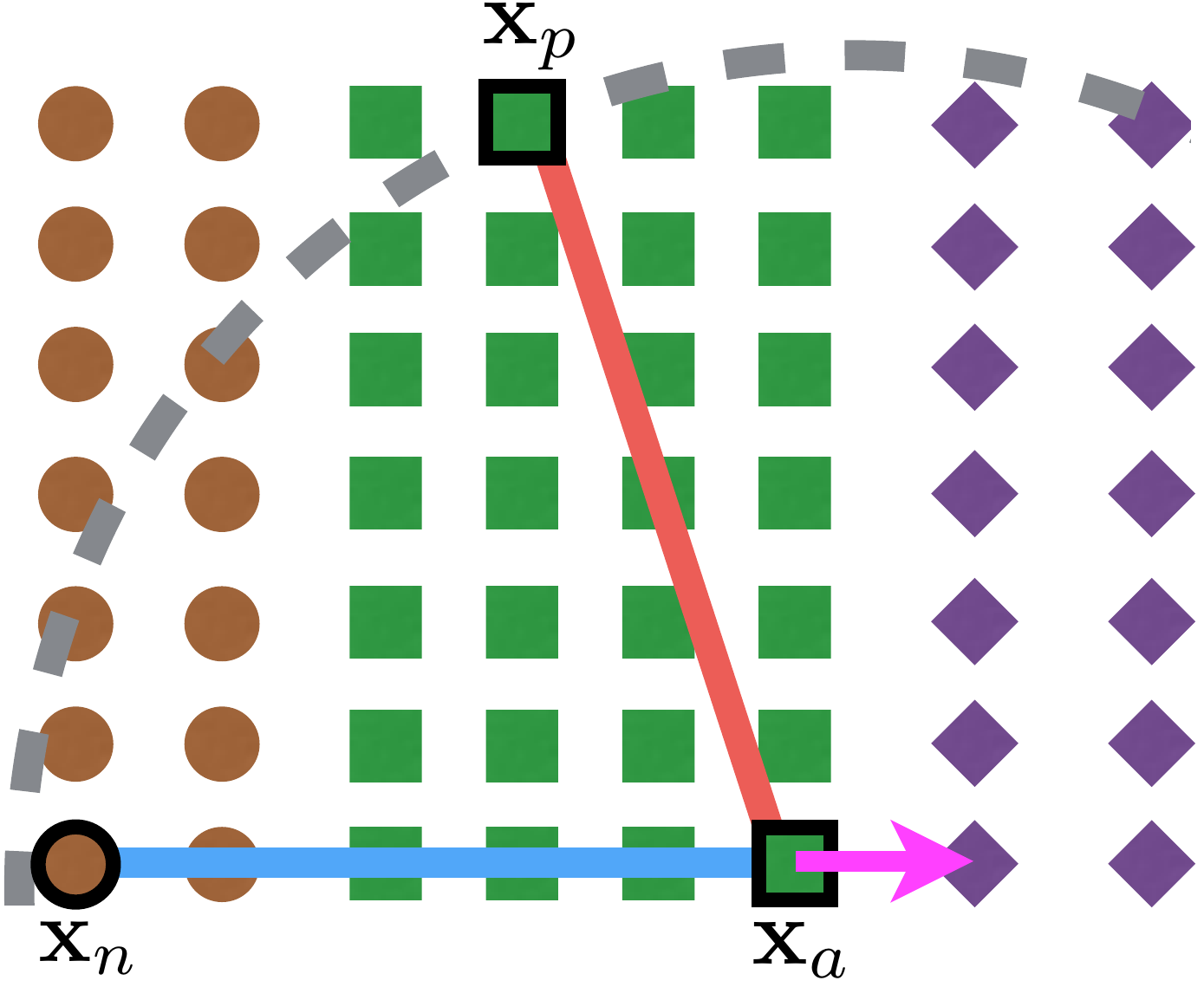} 
\caption{Triplet embedding}
\label{fig:losses_triplet}
\end{subfigure}
\hspace{0.02\textwidth}
\begin{subfigure}[b]{0.33\textwidth}
\includegraphics[width=\textwidth]{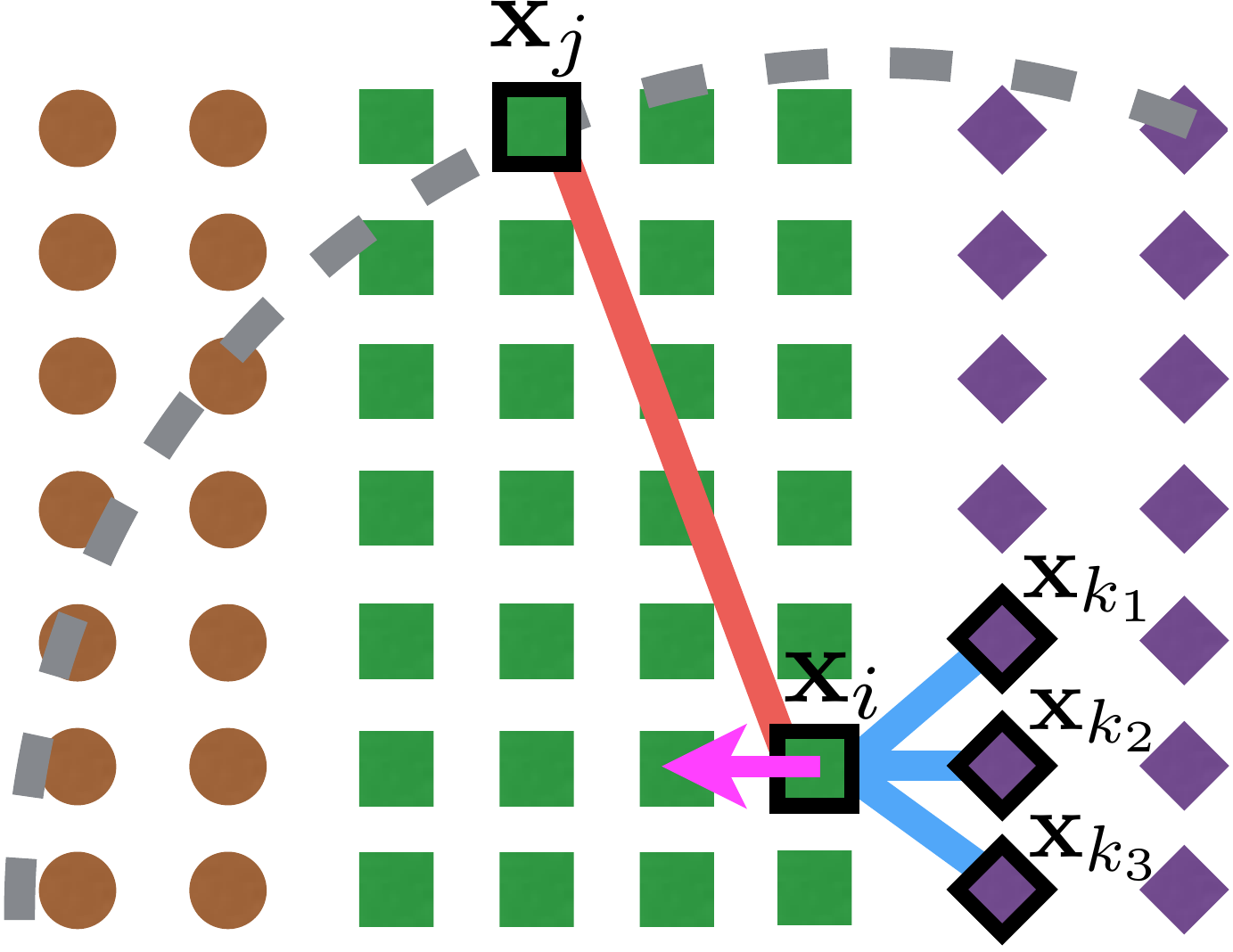}
\caption{Lifted structured similarity}
\label{fig:losses_liftedstructsim}
\end{subfigure}
\caption{Illustration of failure modes of contrastive and triplet loss with randomly sampled training batch. Brown circles, green squares, and purple diamonds represent three different classes. Dotted gray arcs indicate the margin bound (where the loss becomes zero out of the bound) in the hinge loss. Magenta arrows denote the negative gradient direction for the positives.}
\label{fig:losses}
\end{figure*}

\begin{algorithm}
\SetKwInOut{Input}{input}\SetKwInOut{Output}{output}
\Input{~~$D, \alpha$}
\Output{~~$\nicefrac{\partial \tj}{\partial f(\mathbf{x}_i)}, ~~\forall i \in [1, m]$}
\textbf{Initialize: } $\nicefrac{\partial \tj}{\partial f(\mathbf{x}_i)} = \mathbf{0}, ~~\forall i \in [1, m]$\\
\For {$i = 1, \ldots, m$}{
	\For {$j = i+1, \ldots, m, ~~s.t.~ (i,j) \in \mathcal{P}$}{
		\vspace{-0.5em}$$\nicefrac{\partial \tj}{\partial f(\mathbf{x}_i)} \gets \nicefrac{\partial \tj}{\partial f(\mathbf{x}_i)}  +\nicefrac{\partial \tj}{\partial D_{i,j}} \nicefrac{\partial D_{i,j}}{\partial f(\mathbf{x}_i)}$$\vspace{-1.7em}
		$$\nicefrac{\partial \tj}{\partial f(\mathbf{x}_j)} \gets \nicefrac{\partial \tj}{\partial f(\mathbf{x}_j)}  +\nicefrac{\partial \tj}{\partial D_{i,j}} \nicefrac{\partial D_{i,j}}{\partial f(\mathbf{x}_j)}$$
		\For {$k = 1, \ldots, m, ~~ s.t.~ (i,k) \in \mathcal{N}$}{
			\vspace{-0.5em}$$\hspace{-0.4em}\nicefrac{\partial \tj}{\partial f(\mathbf{x}_i)} \gets \nicefrac{\partial \tj}{\partial f(\mathbf{x}_i)}  +\nicefrac{\partial \tj}{\partial D_{i,k}} \nicefrac{\partial D_{i,k}}{\partial f(\mathbf{x}_i)}$$\vspace{-1.7em}
			$$\hspace{-0.4em}\nicefrac{\partial \tj}{\partial f(\mathbf{x}_k)} \gets \nicefrac{\partial \tj}{\partial f(\mathbf{x}_k)}  +\nicefrac{\partial \tj}{\partial D_{i,k}} \nicefrac{\partial D_{i,k}}{\partial f(\mathbf{x}_k)}$$ 
		}
		\For {$l = 1, \ldots, m, ~~ s.t.~ (j,l) \in \mathcal{N}$}{
			\vspace{-0.5em}$$\hspace{-0.4em}\nicefrac{\partial \tj}{\partial f(\mathbf{x}_j)} \gets \nicefrac{\partial \tj}{\partial f(\mathbf{x}_j)}  +\nicefrac{\partial \tj}{\partial D_{j,l}} \nicefrac{\partial D_{j,l}}{\partial f(\mathbf{x}_j)}$$\vspace{-1.7em}
			$$\hspace{-0.4em}\nicefrac{\partial \tj}{\partial f(\mathbf{x}_l)} \gets \nicefrac{\partial \tj}{\partial f(\mathbf{x}_l)}  +\nicefrac{\partial \tj}{\partial D_{j,l}} \nicefrac{\partial D_{j,l}}{\partial f(\mathbf{x}_l)}$$
		}
	}
}
\caption{Backpropagation gradient}
\label{algo:backprop}
\end{algorithm} 

\noindent where $\mathcal{P}$ denotes the set of positive pairs in the batch and $\mathcal{N}$ denotes the set of negative pairs in the batch. The back propagation gradients for the input feature embeddings can be derived as shown in algorithm \ref{algo:backprop}, where the gradients with respect to the distances are,
 
{\small \begin{align}
\frac{\partial \tj}{\partial D_{i,j}} &= \frac{1}{|\mathcal{P}|} \tj_{i,j} ~ \mathbbm{1}{[\tj_{i,j} > 0]}
\label{eqn:dj_pos}
\end{align}}
{\small \begin{align}
\frac{\partial \tj}{\partial D_{i,k}} &= \frac{1}{|\mathcal{P}|} \tj_{i,j} ~ \mathbbm{1}{[\tj_{i,j} > 0]} ~\frac{- \exp\{\alpha - D_{i,k}\}}{\exp\{\tj_{i,j}-D_{i,j}\}}
\label{eqn:dj_neg1}
\end{align}} 
{\small \begin{align}
\frac{\partial \tj}{\partial D_{j,l}} &= \frac{1}{|\mathcal{P}|} \tj_{i,j} ~ \mathbbm{1}{[\tj_{i,j} > 0]} ~\frac{- \exp\{\alpha - D_{j,l}\}}{\exp\{\tj_{i,j}-D_{i,j}\}},
\label{eqn:dj_neg2}
\end{align}}

\noindent where $\mathbbm{1}{[\cdot]}$ is the indicator function which outputs $1$ if the expression evaluates to true and outputs $0$ otherwise. As shown in algorithm \ref{algo:backprop} and equations \ref{eqn:dj_pos}, \ref{eqn:dj_neg1}, and \ref{eqn:dj_neg2}, our method provides informative gradient signals for all negative pairs as long as they are within the margin of any positive pairs (in contrast to only updating the hardest negative) which makes the optimization much more stable. \\

Having stated the formal objective, we now illustrate and discuss some of the failure modes of the contrastive \cite{contrastive} and triplet \cite{facenet, triplet} embedding in which the proposed embedding learns successfully. Figure \ref{fig:losses} illustrates the failure cases in 2D with examples from three different classes. Contrastive embedding (Fig. \ref{fig:losses_contrastive}) can fail if the randomly sampled negative ($\mathbf{x}_j$) is collinear with the examples from another class (purple examples in the figure). Triplet embedding (Fig. \ref{fig:losses_triplet}) can also fail if such sampled negative ($\mathbf{x}_n$) is within the margin bound with respect to the sampled the positive example ($\mathbf{x}_p$) and the anchor ($\mathbf{x}_a$). In this case, both contrastive and triplet embedding incorrectly pushes the positive ($\mathbf{x}_i$/$\mathbf{x}_a$) towards the cluster of examples from the third class. However, in the proposed embedding (Fig. \ref{fig:losses_liftedstructsim}), given sufficiently large random samples $m$, the hard negative examples ($\mathbf{x}_k$'s in Fig. \ref{fig:losses_liftedstructsim}) within the margin bound pushes the positive $\mathbf{x}_i$ towards the correct direction. 

\section{Implementation details}
\label{sec:implementation}

We used the Caffe \cite{jia2014caffe} package for training and testing the embedding with contrastive \cite{contrastive}, triplet \cite{facenet, triplet}, and ours. Maximum training iteration was set to $20,000$ for all the experiments. The margin parameter $\alpha$ was set to $1.0$. The batch size was set to $128$ for contrastive and our method and to $120$ for triplet. For training, all the convolutional layers were initialized from the network pretrained on ImageNet ILSVRC \cite{imagenet} dataset and the fully connected layer (the last layer) was initialized with random weights. We also multiplied the learning rate for the randomly initialized fully connected layers by $10.0$ for faster convergence. All the train and test images are normalized to 256 by 256. For training data augmentation, all images are randomly cropped at 227 by 227 and randomly mirrored horizontally. For training, we exhaustively use all the positive pairs of examples and randomly subsample approximately equal number of negative pairs of examples as positives. 

\begin{figure*}[thbp]
\centering
\includegraphics[width=0.33\textwidth]{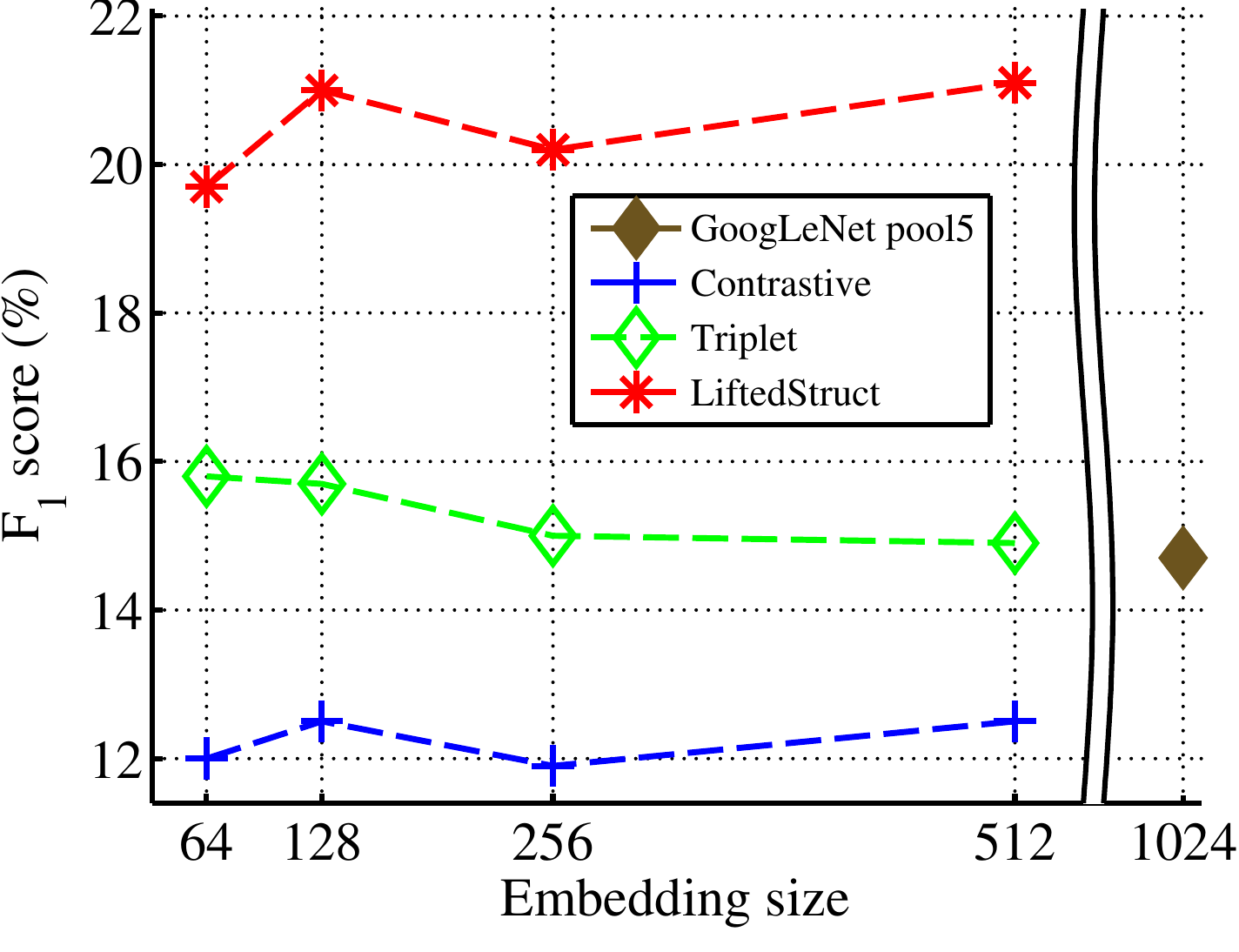}
\includegraphics[width=0.33\textwidth]{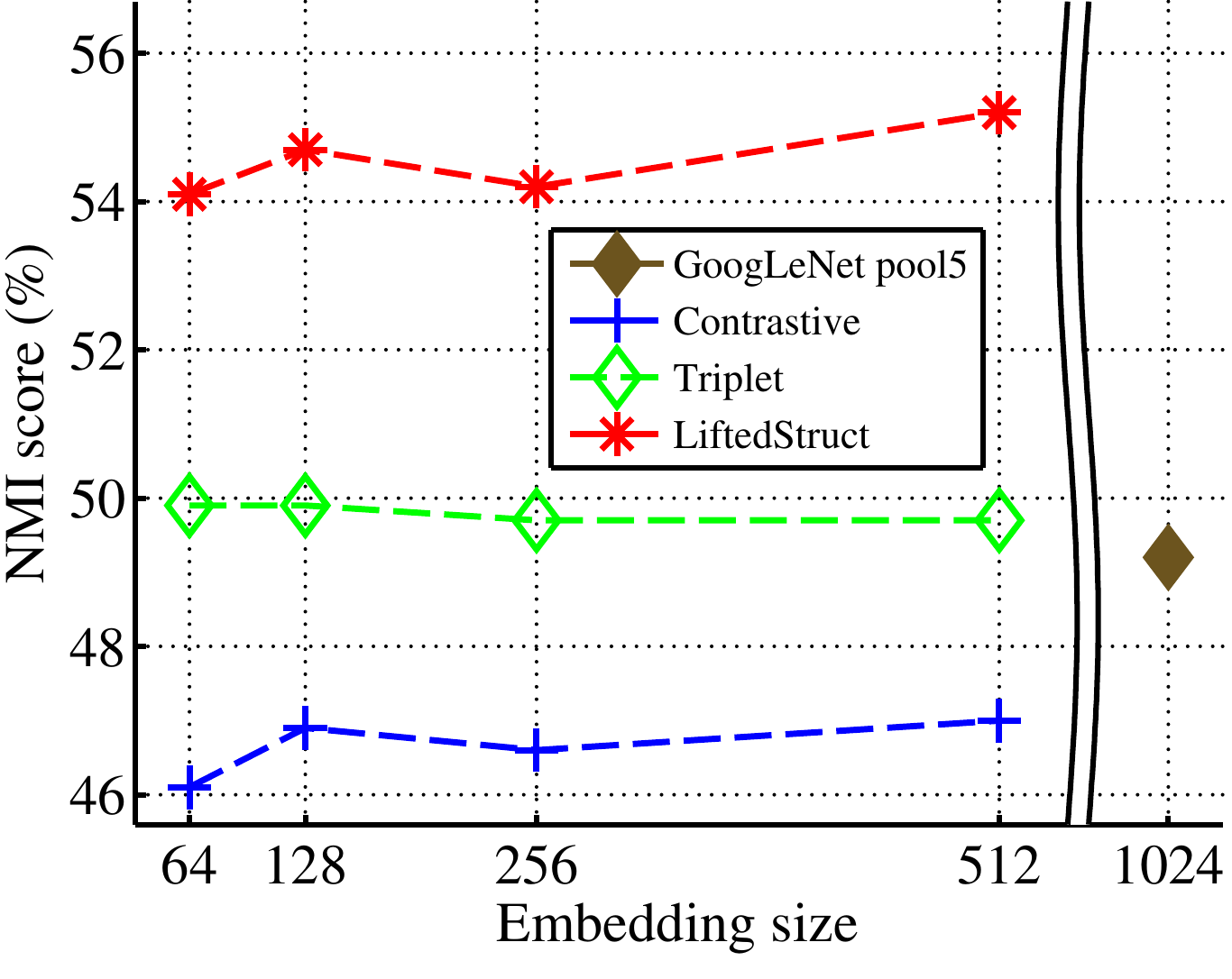}
\includegraphics[width=0.33\textwidth]{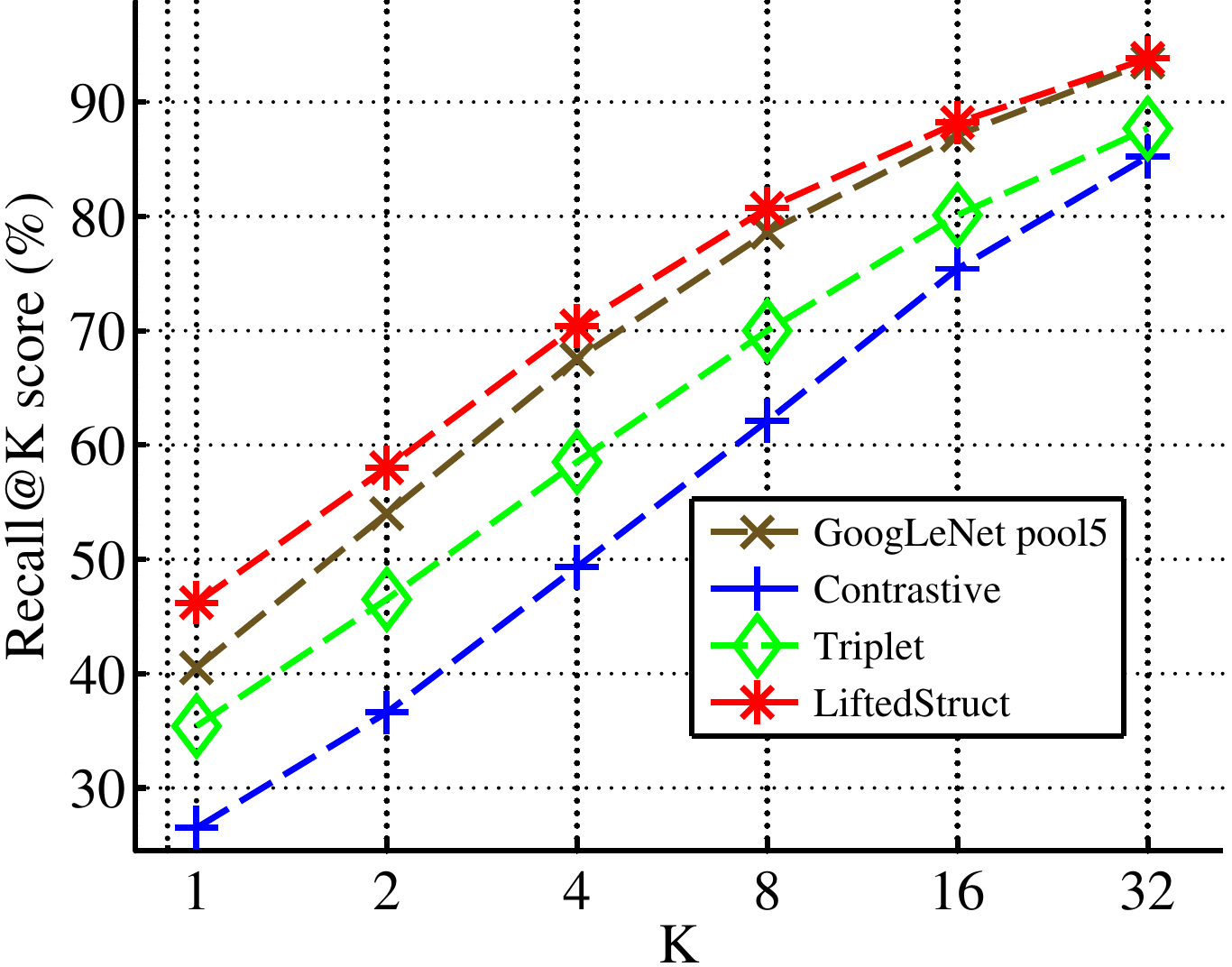}
\caption{$F_1$, NMI, and Recall@K score metrics on the test split of CUB200-2011 with GoogLeNet \cite{googlenet}.}
\label{fig:metric-birds-googlenet}
\end{figure*}

\begin{figure*}[thbp]
\centering
\includegraphics[width=0.33\textwidth]{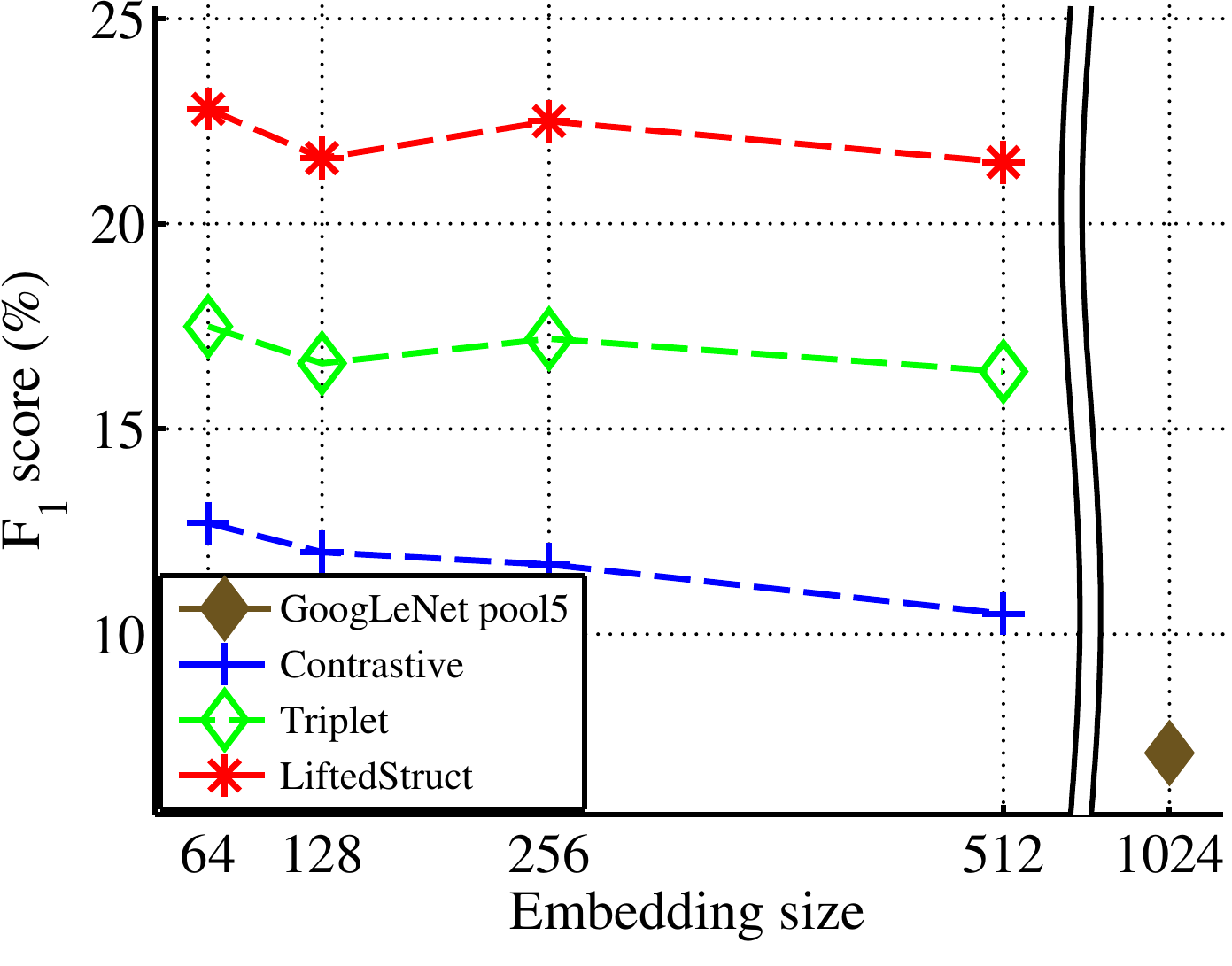}
\includegraphics[width=0.33\textwidth]{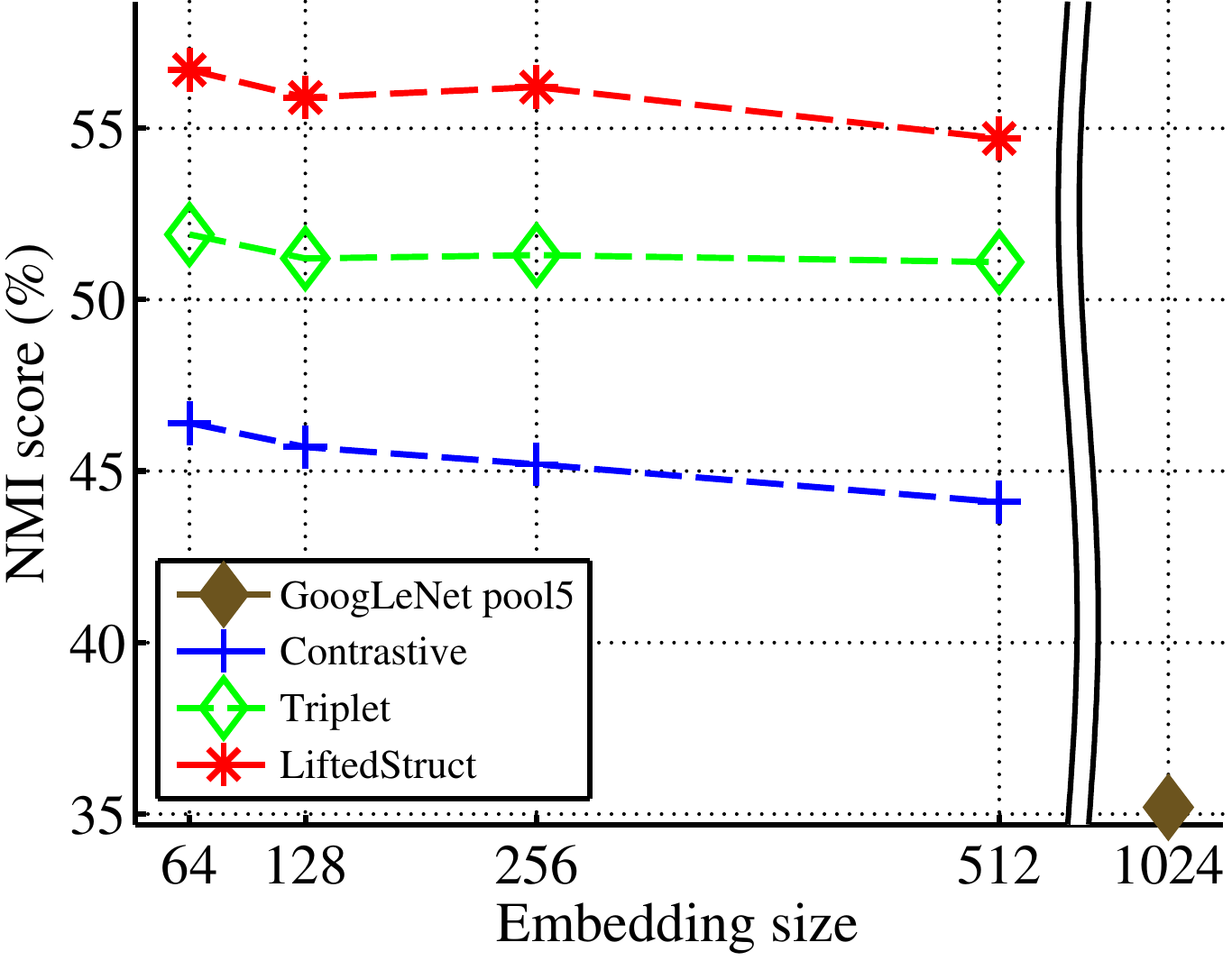}
\includegraphics[width=0.33\textwidth]{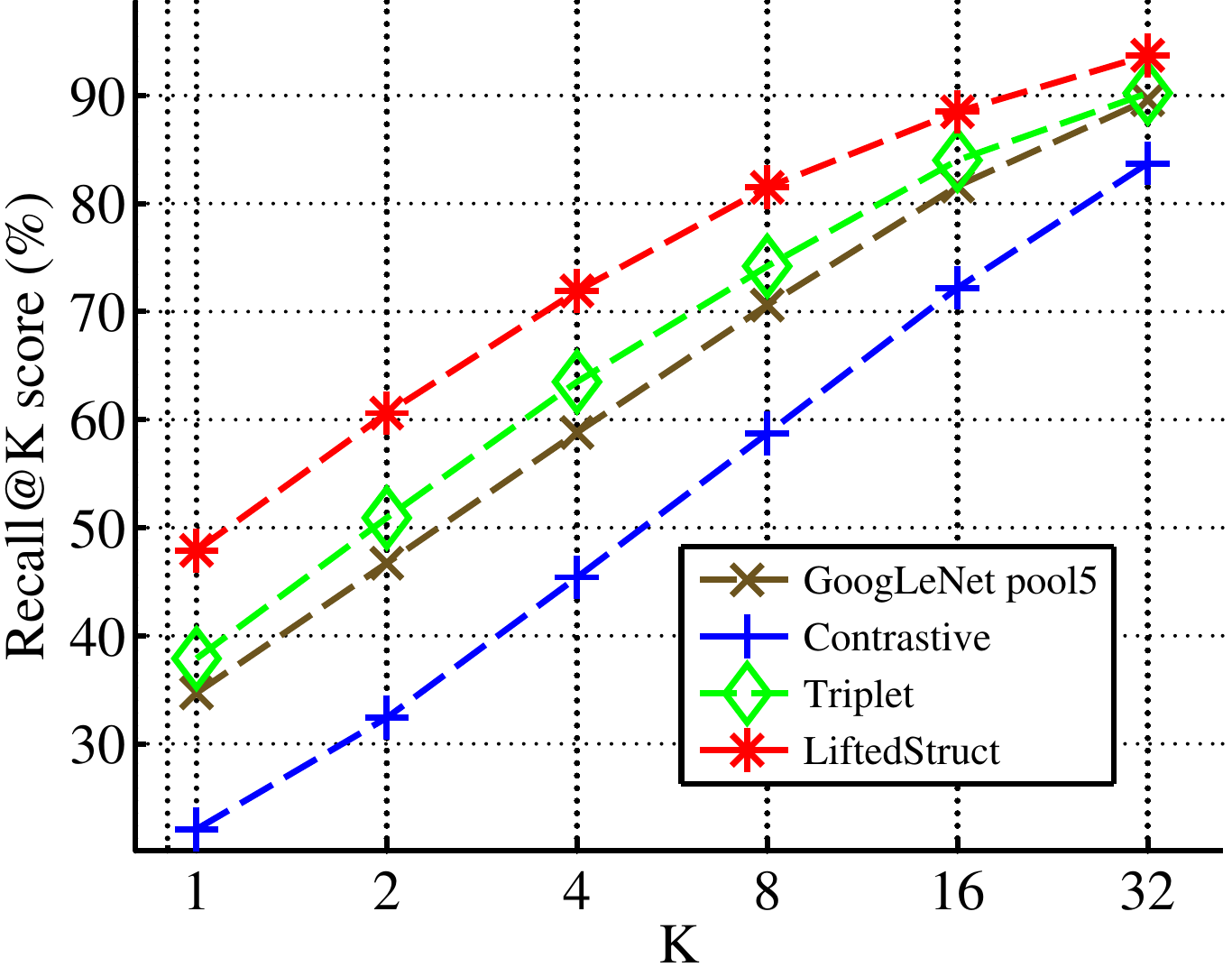}
\caption{$F_1$, NMI, and Recall@K score metrics on the test split of CARS196 with GoogLeNet \cite{googlenet}.}
\label{fig:metric-cars-googlenet}
\end{figure*}

\begin{figure}[thbp]
\centering
\includegraphics[width=0.5\textwidth]{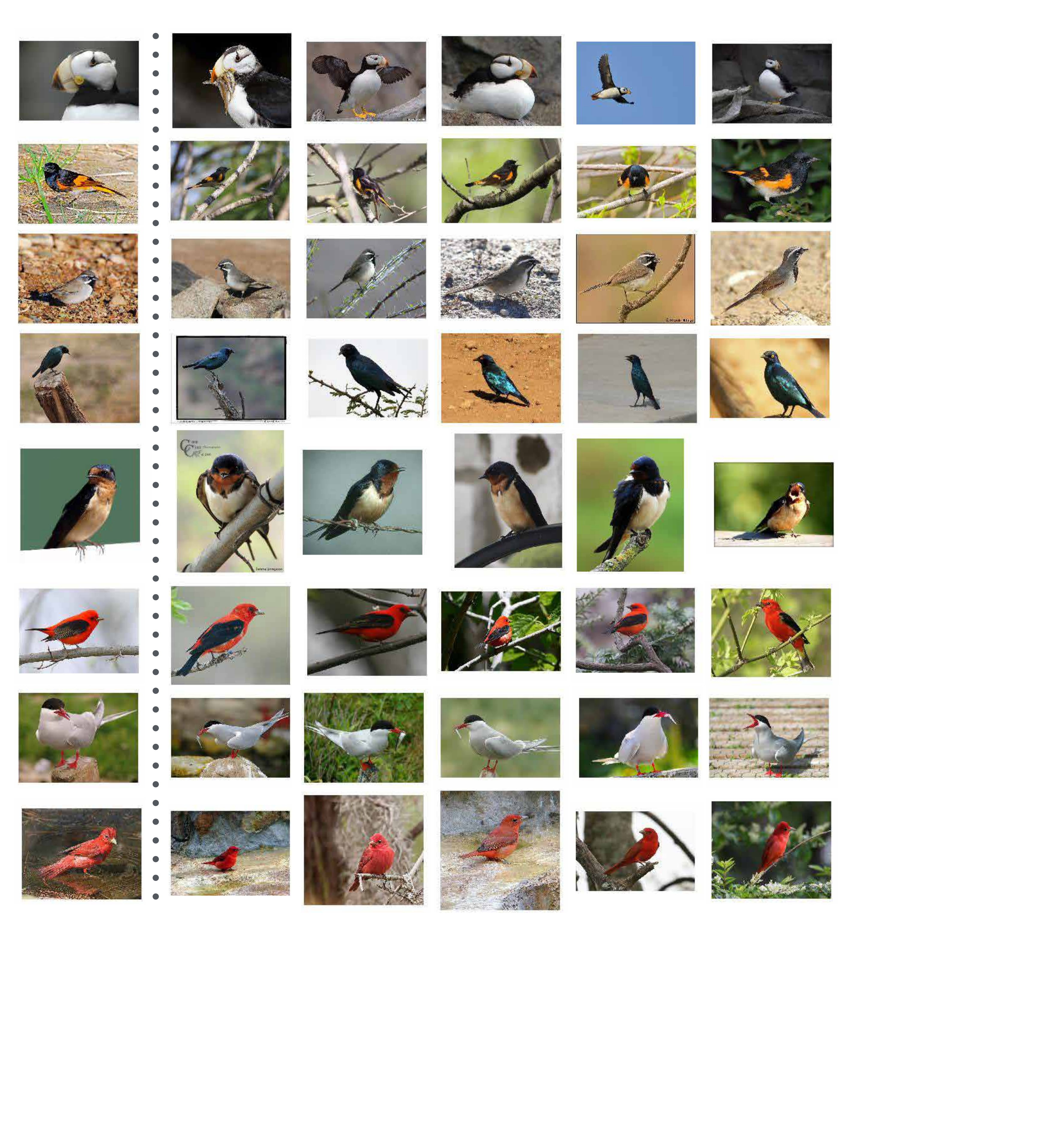}\\
\caption{Examples of successful queries on the CUB200-2011 \cite{cub} dataset using our embedding. Images in the first column are query images and the rest are five nearest neighbors. Best viewed on a monitor zoomed in.}
\label{fig:birds_retrieval_positives}
\end{figure}

\begin{figure*}[thbp]
\centering
\includegraphics[width=0.86\textwidth]{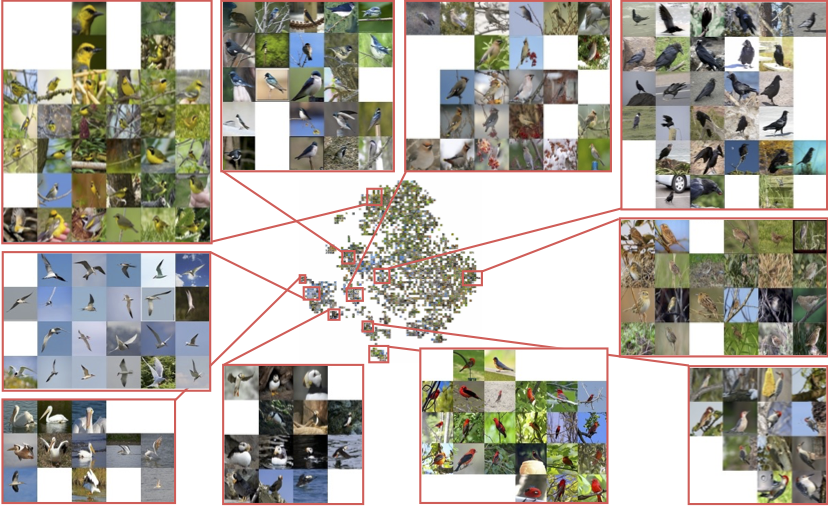}
\caption{Barnes-Hut t-SNE visualization \cite{tsne} of our embedding on the test split (class 101 to 200; 5,924 images) of CUB-200-2011. Best viewed on a monitor when zoomed in.}
\label{fig:tsne-birds}
\end{figure*}

\begin{figure}[thbp]
\centering
\includegraphics[width=0.5\textwidth]{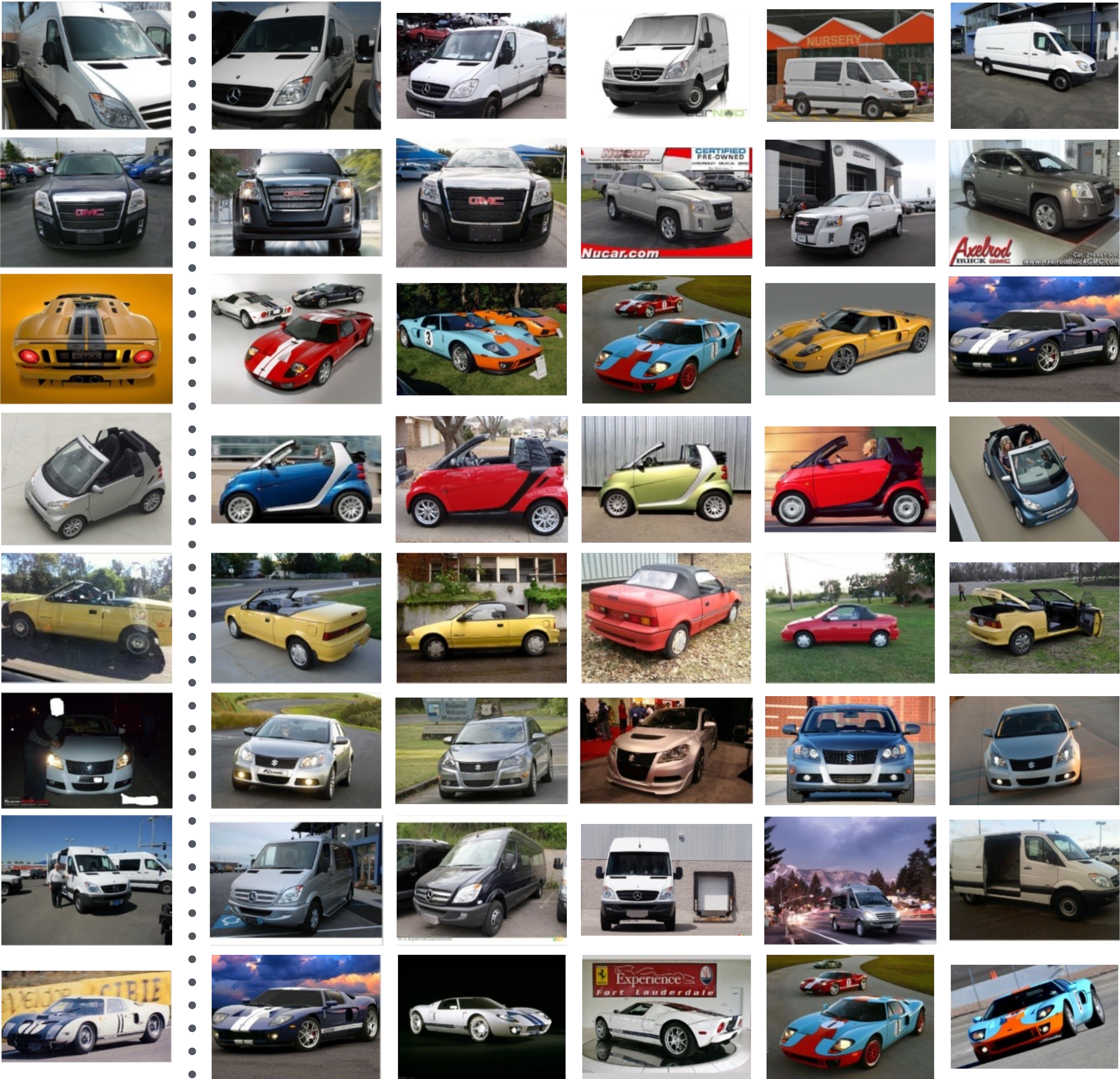}\\
\caption{Examples of successful queries on the Cars196 \cite{cars} dataset using our embedding. Images in the first column are query images and the rest are five nearest neighbors. Best viewed on a monitor zoomed in.}
\label{fig:cars_retrieval_positives}
\end{figure}

\begin{figure*}[thbp]
\centering
\includegraphics[width=0.86\textwidth]{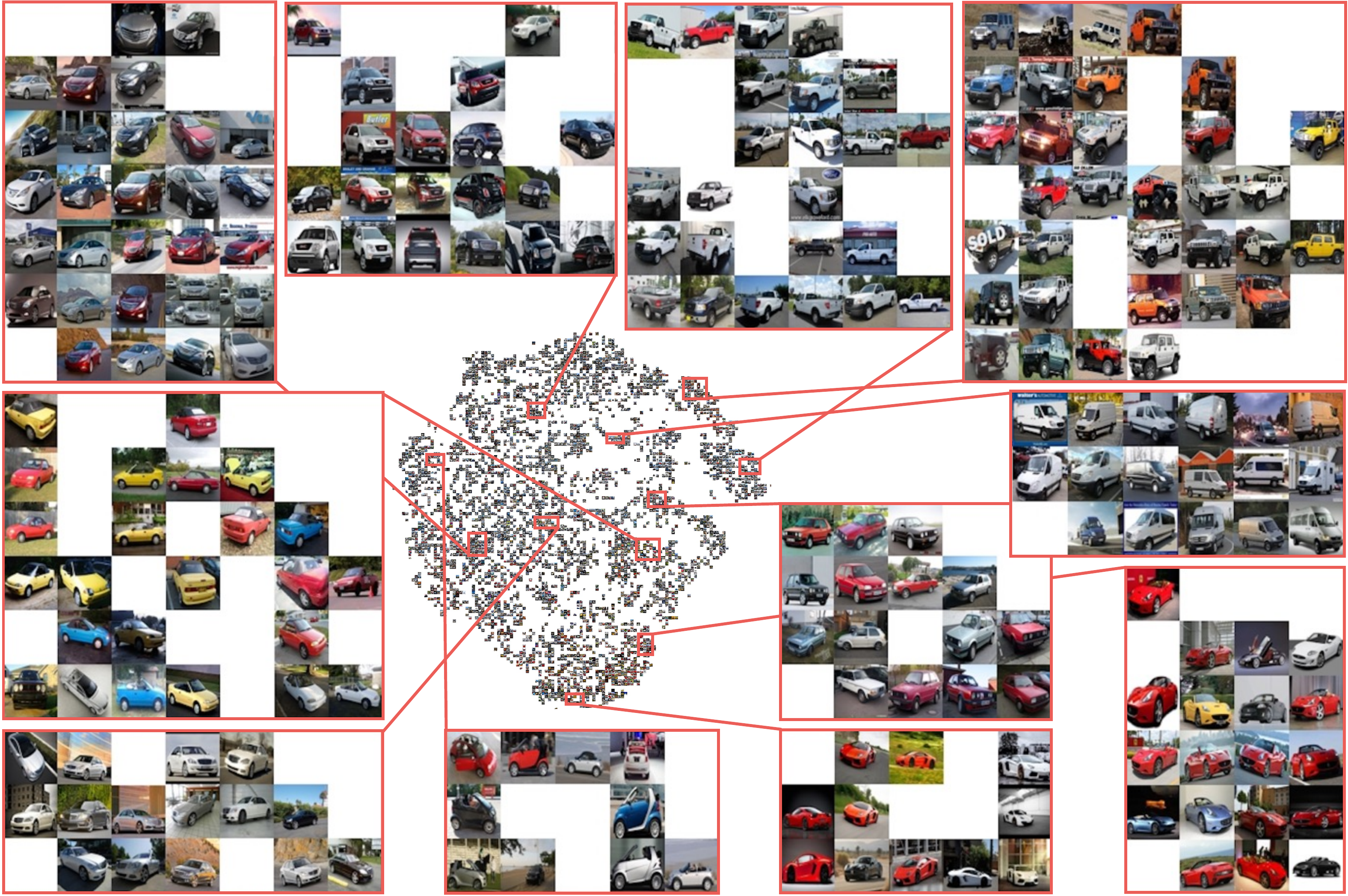}
\caption{Barnes-Hut t-SNE visualization \cite{tsne} of our embedding on the test split (class 99 to 196; 8,131 images) of CARS196. Best viewed on a monitor when zoomed in.}
\label{fig:tsne-cars}
\end{figure*}

\begin{figure*}[thbp]
\centering
\includegraphics[width=0.33\textwidth]{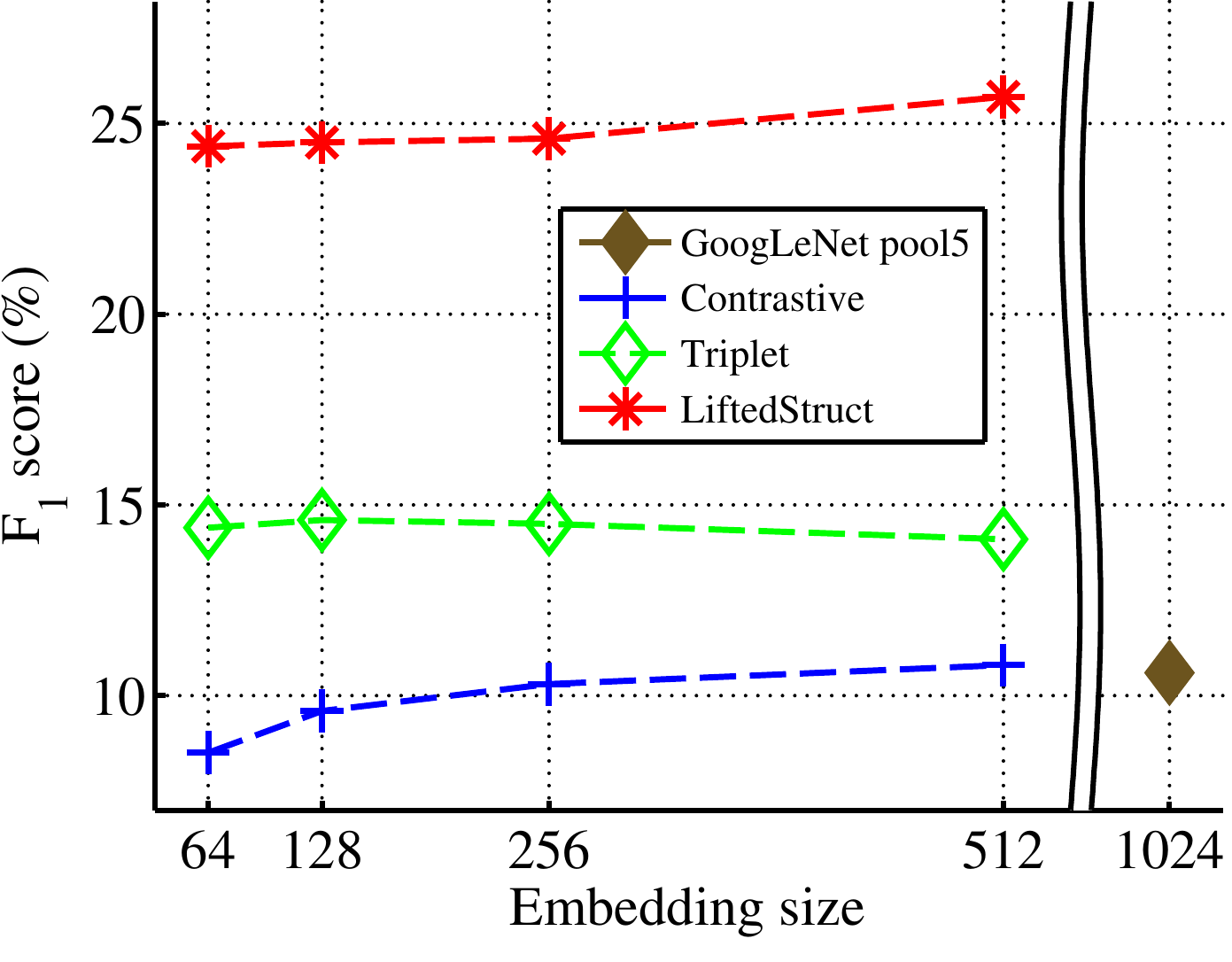}
\includegraphics[width=0.33\textwidth]{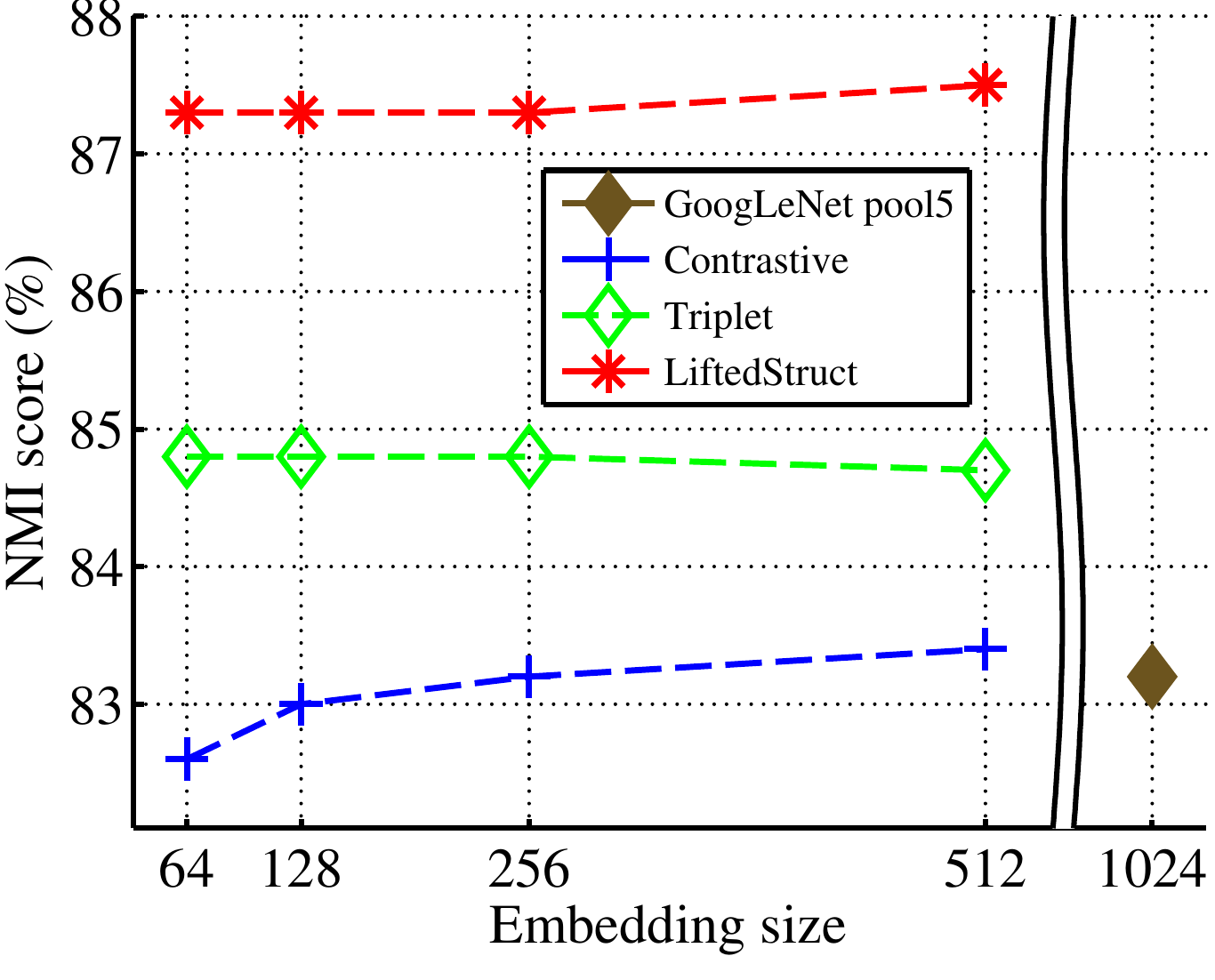}
\includegraphics[width=0.33\textwidth]{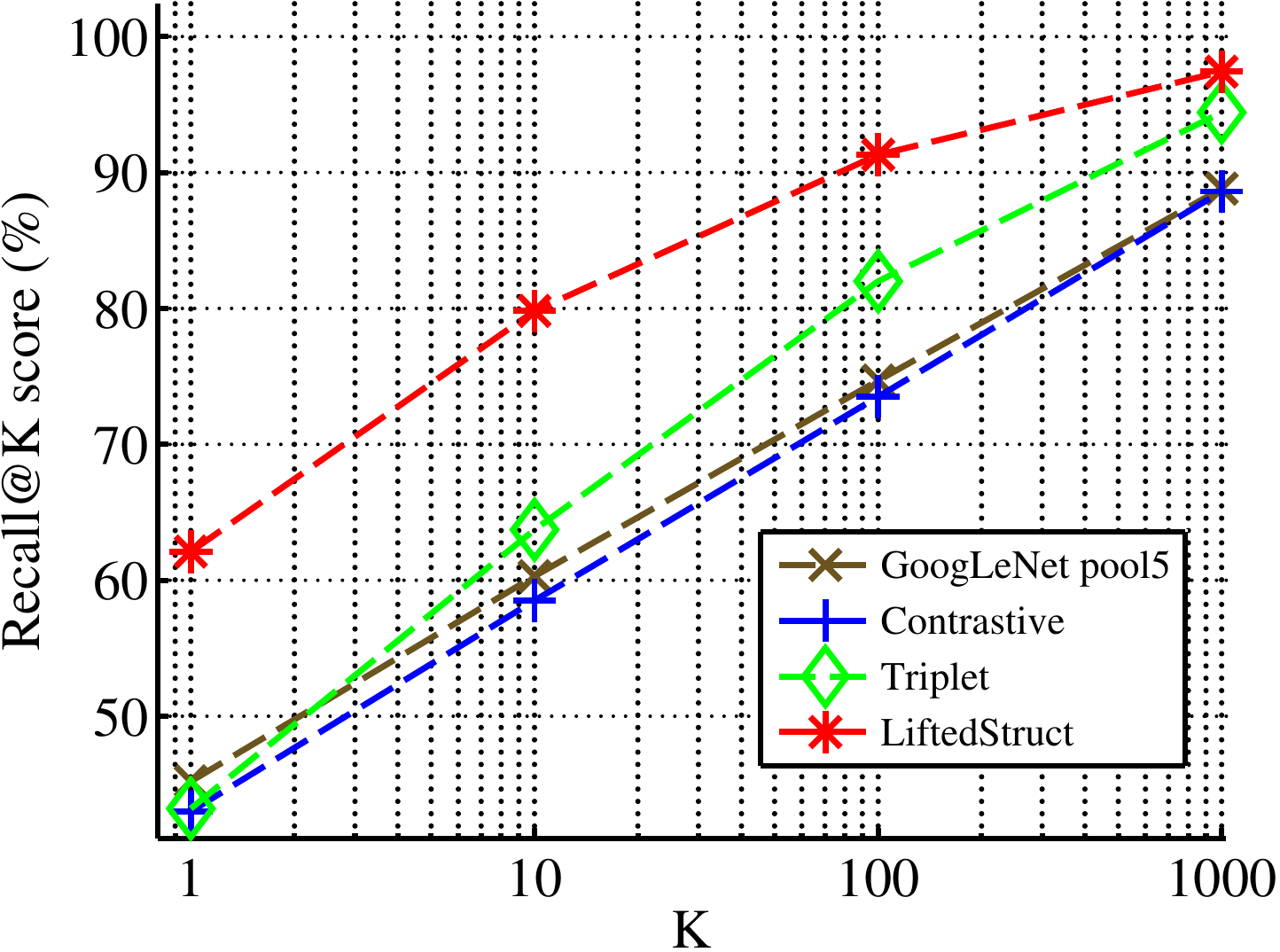}
\caption{$F_1$, NMI, and Recall@K score metrics on the test split of \emph{Online Products} with GoogLeNet \cite{googlenet}.}
\label{fig:recall-products}
\end{figure*}

\begin{figure}[thbp]
\centering
\includegraphics[width=0.5\textwidth]{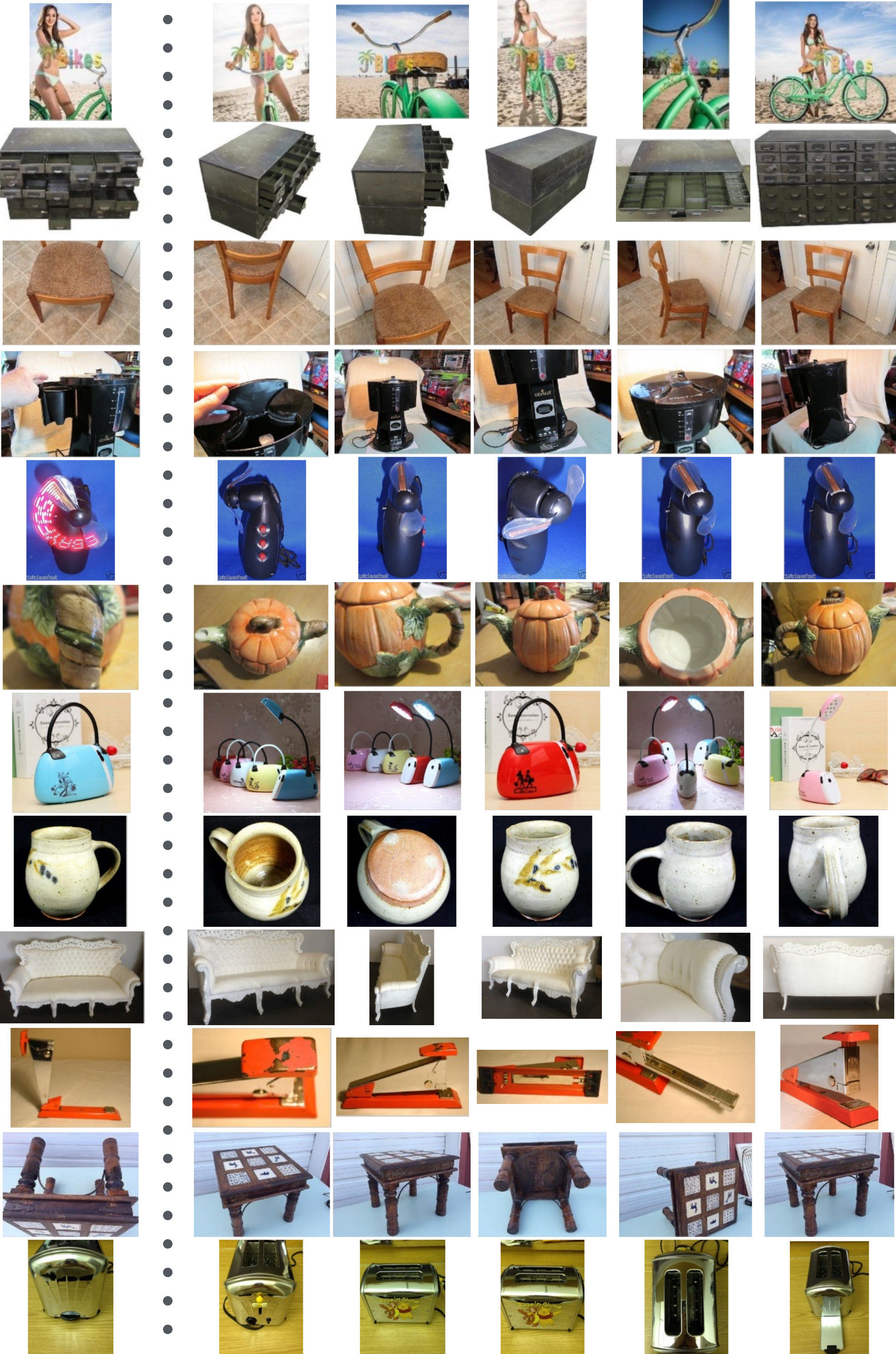}
\caption{Examples of successful queries on our \emph{Online Products} dataset using our embedding (size $512$). Images in the first column are query images and the rest are five nearest neighbors. Best viewed on a monitor zoomed in.}
\label{fig:retrieval_positives}
\end{figure}

\begin{figure}[thbp]
\centering
\includegraphics[width=0.5\textwidth]{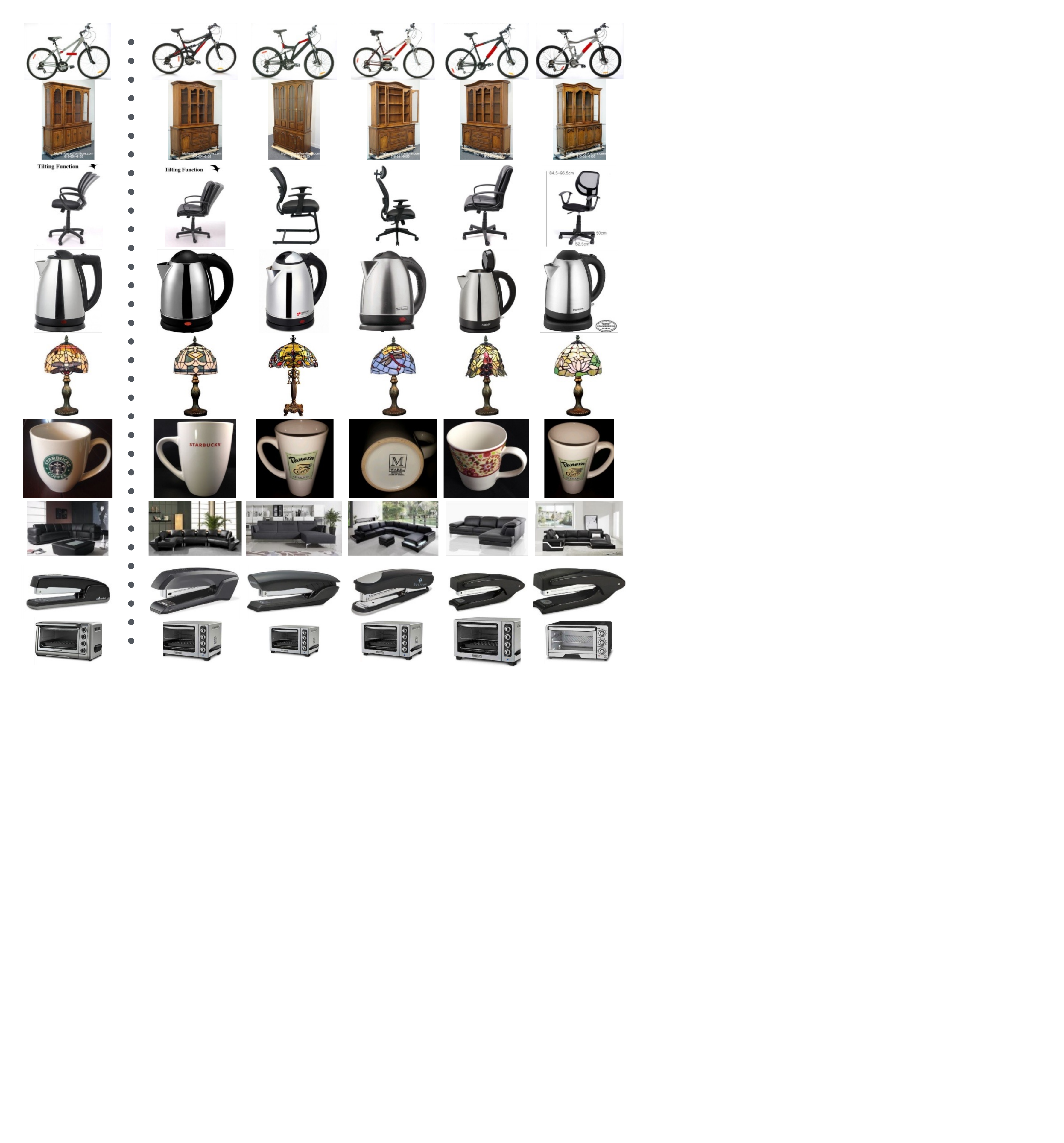}
\caption{Examples of failure queries on \emph{Online Products} dataset. Most failures are fine grained subtle differences among similar products. Images in the first column are query images and the rest are five nearest neighbors. Best viewed on a monitor zoomed in.}
\label{fig:retrieval_negatives}
\end{figure}

\begin{figure*}[thbp]
\centering
\includegraphics[width=\textwidth]{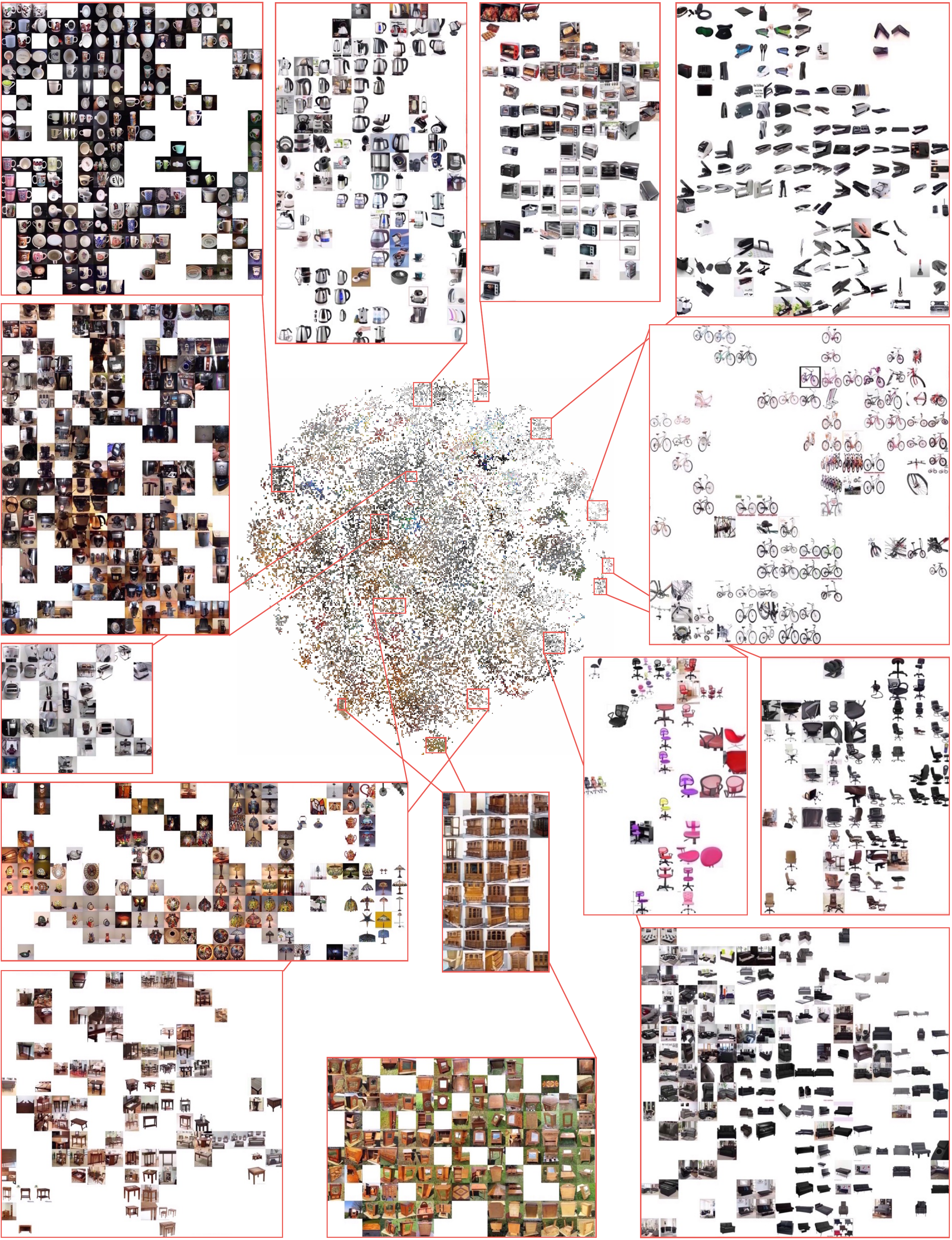} \vspace{-2.2em}
\caption{Barnes-Hut t-SNE visualization \cite{tsne} of our embedding on the test split (class 11,319 to 22,634; 60,502 images) of \emph{Online Products}.} 
\label{fig:ebay_tsne}
\end{figure*}

\section{Evaluation}
\label{sec:evaluation}


In this section, we briefly introduce the evaluation metrics used in the experiments. For the clustering task, we use the $\text{F}_1$ and NMI metrics. $\text{F}_1$ metric computes the harmonic mean of precision and recall. $\text{F}_1 = \frac{2 P R}{P + R}$. The normalized mutual information (NMI) metric take as input a set of clusters $\Omega = \{ \omega_1, \ldots, \omega_K \}$ and a set of ground truth classes $\mathbb{C} = \{ c_1, \ldots, c_K  \}$. $\omega_i$ indicates the set of examples with cluster assignment $i$. $c_j$ indicates the set of examples with the ground truth class label $j$. Normalized mutual information is defined by the ratio of mutual information and the average entropy of clusters and the entropy of labels. $\text{NMI}\left(\Omega, \mathbb{C}\right) = \frac{I\left(\Omega; \mathbb{C}\right)}{2\left(H\left(\Omega\right) + H\left(\mathbb{C}\right)\right)}$. We direct interested readers to refer  \cite{manningbook} for complete details. For the retrieval task, we use the Recall@K \cite{recall_at_K} metric. Each test image (query) first retrieves K nearest neighbors from the test set and receives score $1$ if an image of the same class is retrieved among the K nearest neighbors and $0$ otherwise. Recall@K averages this score over all the images. 

\section{Experiments}
\label{sec:experiments}

We show experiments on CUB200-2011 \cite{cub}, CARS196 \cite{cars}, and our \emph{Online Products} datasets where we use the first half of classes for training and the rest half classes for testing. For testing, we first compute the embedding on all the test images at varying embedding sizes $\{64, 128, 256, 512\}$ following the practice in \cite{seanbell, facenet}. For clustering evaluation, we run affinity propagation clustering \cite{affinityprop} with bisection method \cite{apclusterK} for the desired number of clusters set equal to the number of classes in the test set. The clustering quality is measured in the standard $F_1$ and NMI metrics. For the retrieval evaluation, we report the result on the standard Recall@K metric \cite{recall_at_K} in log space of K. The experiments are performed with GoogLeNet \cite{googlenet}. 

\subsection{CUB-200-2011}

The CUB-200-2011 dataset \cite{cub} has 200 classes of birds with 11,788 images. We split the first $100$ classes for training (5,864 images) and the rest of the classes for testing (5,924 images). Figure \ref{fig:metric-birds-googlenet} shows the quantitative clustering quality for the contrastive \cite{contrastive}, triplet \cite{triplet, facenet}, and using pool5 activation from the pretrained GoogLeNet \cite{googlenet} network on ImageNet \cite{imagenet}, and our method on both $\text{F}_1$, NMI, and Recall@K metrics. Our embedding shows significant performance margin both on the standard $\text{F}_1$, NMI, and Recall@K metrics on all the embedding sizes. Figure \ref{fig:birds_retrieval_positives} shows some example query and nearest neighbors on the test split of CUB200-2011 \cite{cub} dataset. Figure \ref{fig:tsne-birds} shows the Barnes-Hut t-SNE visualization \cite{tsne} on our $64$ dimensional embedding. Although t-SNE embedding does not directly translate to the high dimensional embedding, it is clear that similar types of birds are quite clustered together and are apart from other species.

\subsection{CARS196}
The CARS196 data set \cite{cars} has 198 classes of cars with 16,185 images. We split the first 98 classes for training (8,054 images) and the other 98 classes for testing (8,131 images). Figure \ref{fig:metric-cars-googlenet} shows the quantitative clustering quality for the contrastive \cite{contrastive}, triplet \cite{triplet, facenet}, and using pool5 activation from pretrained GoogLeNet \cite{googlenet} network on ImageNet \cite{imagenet}. Our embedding shows significant margin in terms of the standard $\text{F}_1$, NMI, and Recall@K metrics on all the embedding sizes. Figure \ref{fig:cars_retrieval_positives} shows some example query and nearest neighbors on the test split of Cars196 \cite{cars} dataset. Figure \ref{fig:tsne-cars} shows the Barnes-Hut t-SNE visualization \cite{tsne} on our $64$ dimensional embedding. We can observe that the embedding clusters the images from the same brand of cars despite the significant pose variations and the changes in the body paint. 

\subsection{Online Products dataset}
We used the web crawling API from eBay.com \cite{ebay_api} to download images and filtered duplicate and irrelevant images (i.e. photos of contact phone numbers, logos, etc). The preprocessed dataset has 120,053 images of 22,634 online products (classes) from eBay.com. Each product has approximately $5.3$ images. For the experiments, we split 59,551 images of 11,318 classes for training and 60,502 images of 11,316 classes for testing. Figure \ref{fig:recall-products} shows the quantitative clustering and retrieval results  on $\text{F}_1$, NMI, and Recall@K metric with GoogLeNet. Figures \ref{fig:retrieval_positives} and \ref{fig:retrieval_negatives} show some example queries and nearest neighbors on the dataset for both successful and failure cases. Despite the huge changes in the viewpoint, configuration, and illumination, our method can successfully retrieve examples from the same class and most retrieval failures come from fine grained subtle differences among similar products. Figure \ref{fig:ebay_tsne} shows the t-SNE visualization of the learned embedding on our \emph{Online Products} dataset.

\section{Conclusion}

We described a deep feature embedding and metric learning algorithm which defines a novel structured prediction objective on the lifted dense pairwise distance matrix within the batch during the neural network training. The experimental results on CUB-200-2011 \cite{cub}, CARS196 \cite{cars}, and our \emph{Online Products} datasets show state of the art performance on all the experimented embedding dimensions. 

{\small
\bibliographystyle{ieee}
\bibliography{mainbib}

\begin{thebibliography}{10}\itemsep=-1pt

\bibitem{seanbell}
S.~Bell and K.~Bala.
\newblock Learning visual similarity for product design with convolutional
  neural networks.
\newblock In {\em SIGGRAPH}, 2015.

\bibitem{outofsample}
Y.~Bengio, J.~Paiement, and P.~Vincent.
\newblock Out-of-sample extensions for lle, isomap, mds, eigenmaps, and
  spectral clustering.
\newblock In {\em NIPS}, 2004.

\bibitem{signatureVerification}
J.~Bromley, I.~Guyon, Y.~Lecun, E.~SŠckinger, and R.~Shah.
\newblock Signature verification using a ``siamese" time delay neural network.
\newblock In {\em NIPS}, 1994.

\bibitem{triplet_ranking}
G.~Chechik, V.~Sharma, U.~Shalit, and S.~Bengio.
\newblock Large scale online learning of image similarity through ranking.
\newblock {\em JMLR}, 11, 2010.

\bibitem{faceVerification}
S.~Chopra, R.~Hadsell, and Y.~LeCun.
\newblock Learning a similarity metric discriminatively, with application to
  face verification.
\newblock In {\em CVPR}, June 2005.

\bibitem{extreme_langford}
A.~Choromanska, A.~Agarwal, and J.~Langford.
\newblock Extreme multi class classification.
\newblock In {\em NIPS}, 2013.

\bibitem{mds}
T.~Cox and M.~Cox.
\newblock Multidimensional scaling.
\newblock In {\em London: Chapman and Hill}, 1994.

\bibitem{triplet_data}
I.~S. Data.
\newblock \url{https://sites.google.com/site/imagesimilaritydata/}, 2014.

\bibitem{ebay_api}
eBay Developers~Program.
\newblock \url{http://go.developer.ebay.com/what-ebay-api}, 2015.

\bibitem{apclusterK}
B.~J. Frey and D.~Dueck.
\newblock apclusterk.m.
\newblock \url{http://www.psi.toronto.edu/affinitypropagation/apclusterK.m},
  2007.

\bibitem{affinityprop}
B.~J. Frey and D.~Dueck.
\newblock Clustering by passing messages between data points.
\newblock {\em Science}, 2007.

\bibitem{devise}
A.~Frome, G.~S. Corrado, J.~Shlens, S.~Bengio, J.~Dean, M.~Ranzato, and
  T.~Mikolov.
\newblock Devise: A deep visual-semantic embedding model.
\newblock In {\em NIPS}, 2013.

\bibitem{nca}
J.~Goldberger, S.~Roweis, G.~Hinton, and R.~Salakhutdinov.
\newblock Neighbourhood component analysis.
\newblock In {\em NIPS}, 2004.

\bibitem{contrastive}
R.~Hadsell, S.~Chopra, and Y.~Lecun.
\newblock Dimensionality reduction by learning an invariant mapping.
\newblock In {\em CVPR}, 2006.

\bibitem{recall_at_K}
H.~Jegou, M.~Douze, and C.~Schmid.
\newblock Product quantization for nearest neighbor search.
\newblock In {\em PAMI}, 2011.

\bibitem{jia2014caffe}
Y.~Jia, E.~Shelhamer, J.~Donahue, S.~Karayev, J.~Long, R.~Girshick,
  S.~Guadarrama, and T.~Darrell.
\newblock Caffe: Convolutional architecture for fast feature embedding.
\newblock {\em arXiv preprint arXiv:1408.5093}, 2014.

\bibitem{Joachims/etal/09a}
T.~Joachims, T.~Finley, and C.-N. Yu.
\newblock Cutting-plane training of structural svms.
\newblock {\em JMLR}, 2009.

\bibitem{pca}
T.~I. Jolliffe.
\newblock Principal component analysis.
\newblock In {\em New York: Springer-Verlag}, 1986.

\bibitem{cars}
J.~Krause, M.~Stark, J.~Deng, and F.-F. Li.
\newblock 3d object representations for fine-grained categorization.
\newblock {\em ICCV 3dRR-13}, 2013.

\bibitem{alexnet}
A.~Krizhevsky, I.~Sutskever, and G.~Hinton.
\newblock Imagenet classification with deep convolutional neural networks.
\newblock In {\em NIPS}, 2012.

\bibitem{lampert}
C.~H. Lampert, H.~Nickisch, and S.~Harmeling.
\newblock Attribute-based classification for zero-shot visual object
  categorization.
\newblock In {\em TPAMI}, 2014.

\bibitem{chairs}
Y.~Li, H.~Su, C.~Qi, N.~Fish, D.~Cohen-Or, and L.~Guibas.
\newblock Joint embeddings of shapes and images via cnn image purification.
\newblock In {\em SIGGRAPH Asia}, 2015.

\bibitem{manningbook}
C.~D. Manning, P.~Raghavan, and H.~SchŸtze.
\newblock {\em Introduction to Information Retrieval}.
\newblock Cambridge university press, 2008.

\bibitem{mensink}
T.~Mensink, J.~Verbeek, F.~Perronnin, and G.~Csurk.
\newblock Metric learning for large scale image classification: Generalizaing
  to new classes at near-zero cost.
\newblock In {\em ECCV}, 2012.

\bibitem{palatucci}
M.~Palatucci, D.~Pomerleau, G.~E. Hinton, and T.~M. Mitchell.
\newblock Zero-shot learning with semantic output codes.
\newblock In {\em NIPS}, 2009.

\bibitem{extreme_varma}
Y.~Prabhu and M.~Varma.
\newblock Fastxml: A fast, accurate and stable tree-classifier for extreme
  multi-label learning.
\newblock In {\em SIGKDD}, 2014.

\bibitem{nec}
Q.~Qian, R.~Jin, S.~Zhu, and Y.~Lin.
\newblock Fine-grained visual categorization via multi-stage metric learning.
\newblock In {\em CVPR}, 2015.

\bibitem{rohrbach}
M.~Rohrbach, M.~Stark, and B.~Schiel.
\newblock Evaluating knowledge transfer and zero-shot learn- ing in a
  large-scale setting.
\newblock In {\em CVPR}, 2011.

\bibitem{lle}
S.~Roweis and L.~Saul.
\newblock Nonlinear dimensionality reduction by locally linear embedding.
\newblock In {\em Science}, 290.

\bibitem{imagenet}
O.~Russakovsky, J.~Deng, H.~Su, J.~Krause, S.~Satheesh, S.~Ma, Z.~Huang,
  A.~Karpathy, A.~Khosla, M.~Bernstein, A.~C. Berg, and L.~Fei-Fei.
\newblock {ImageNet Large Scale Visual Recognition Challenge}.
\newblock {\em IJCV}, 2015.

\bibitem{facenet}
F.~Schroff, D.~Kalenichenko, and J.~Philbin.
\newblock Facenet: A unified embedding for face recognition and clustering.
\newblock In {\em CVPR}, 2015.

\bibitem{socher}
R.~Socher, C.~D.~M. M.~Ganjoo H.~Sridhar, O.~Bastani, and A.~Y. Ng.
\newblock Zero-shot learning through cross-modal transfer.
\newblock In {\em ICLR}, 2013.

\bibitem{googlenet}
C.~Szegedy, W.~Liu, Y.~Jia, P.~Sermanet, S.~Reed, D.~Anguelov, D.~Erhan,
  V.~Vanhoucke, and A.~Rabinovich.
\newblock Going deeper with convolutions.
\newblock In {\em CVPR}, 2015.

\bibitem{taylor}
G.~Taylor, R.~Fergus, G.~Williams, I.~Spiro, and C.~Bregler.
\newblock Pose-sensitive embedding by nonlinear nca regression.
\newblock In {\em NIPS}, 2010.

\bibitem{Tsochantaridis/etal/04}
I.~Tsochantaridis, T.~Hofmann, T.~Joachims, and Y.~Altun.
\newblock Support vector machine learning for interdependent and structured
  output spaces.
\newblock In {\em ICML}, 2004.

\bibitem{tsne}
L.~van~der maaten.
\newblock Accelerating t-sne using tree-based algorithms.
\newblock In {\em JMLR}, 2014.

\bibitem{cub}
C.~Wah, S.~Branson, P.~Welinder, P.~Perona, and S.~Belongie.
\newblock The caltech-ucsd birds-200-2011 dataset.
\newblock Technical Report CNS-TR-2011-001, California Institute of Technology,
  2011.

\bibitem{triplet_cvpr14}
J.~Wang, Y.~Song, T.~Leung, C.~Rosenberg, J.~Wang, J.~Philbin, B.~Chen, and
  Y.~Wu.
\newblock Learning fine-grained image similarity with deep ranking.
\newblock In {\em CVPR}, 2014.

\bibitem{triplet}
K.~Q. Weinberger, J.~Blitzer, and L.~K. Saul.
\newblock Distance metric learning for large margin nearest neighbor
  classification.
\newblock In {\em NIPS}, 2006.

\bibitem{wasabi}
J.~Weston, S.~Bengio, and N.~Usurer.
\newblock Wsabi: Scaling up to large vocabulary image annotation.
\newblock In {\em IJCAI}, 2011.

\end{thebibliography}
}

\end{document}